\documentclass[sigconf,nonacm]{acmart}
\usepackage{extarrows}
\usepackage{color} 
\usepackage{xcolor}
\usepackage{makecell}
\usepackage{amsmath,amsfonts}
\usepackage{algorithmic}
\usepackage{array}
\usepackage{graphicx}
\usepackage{stfloats}
\usepackage{multicol}
\usepackage{multirow}
\usepackage{hhline}
\usepackage{colortbl}
\usepackage{bm}
\usepackage{url}
\usepackage{enumitem}
\usepackage{CJKutf8}
\usepackage{tabularray}
\usepackage{bigstrut}
\usepackage{booktabs}
\usepackage{arydshln}
\usepackage{utfsym}
\usepackage{hyperref}
\usepackage{float}
\usepackage{threeparttable}
\usepackage{bibunits}

\definecolor{lightgray}{rgb}{0.88, 0.92, 0.98}
\definecolor{defblue}{rgb}{0.1843, 0.3333, 0.6}
\definecolor{defred}{rgb}{0.88, 0.2510, 0.3294}

\definecolor{defgreen1}{rgb}{ 0.910,  0.953,  0.855}
\definecolor{defgreen2}{rgb}{0.82,  0.902,  0.710}
\definecolor{defgreen3}{rgb}{0.713,  0.903,  0.648}
\definecolor{defgreen4}{rgb}{ 0.725,  0.855,  0.561}

\definecolor{defyellow}{rgb}{1,  0.983,  0.717}
\definecolor{defyellowtext}{rgb}{1,  0.851,  0.438}
\definecolor{defred3}{rgb}{ 1,  0.398,  0.399}

\definecolor{customgreen}{rgb}{0.3647, 0.6784, 0.3294}

\AtBeginDocument{%
  \providecommand\BibTeX{{%
    \normalfont B\kern-0.5em{\scshape i\kern-0.25em b}\kern-0.8em\TeX}}}

\copyrightyear{2025}
\acmYear{2025}
\setcopyright{acmlicensed}\acmConference[MM '25]{Proceedings of the 33rd ACM International Conference on Multimedia}{October 27--31, 2025}{Dublin, Ireland}
\acmBooktitle{Proceedings of the 33rd ACM International Conference on Multimedia (MM '25), October 27--31, 2025, Dublin, Ireland}
\acmDOI{xxx}
\acmISBN{xxx}

\begin{document}
% \begin{CJK}{UTF8}{gbsn}
% \begin{sloppypar}
\title{HUD: Hierarchical Uncertainty-Aware Disambiguation Network for Composed Video Retrieval}

\author{Zhiwei Chen}
\orcid{0009-0003-0365-8553}
\affiliation{
\department{\normalsize School of Software}
  \institution{\normalsize Shandong University}
  \city{Jinan}
  \country{China}
  }
\email{zivczw@gmail.com}

\author{Yupeng Hu}
\orcid{0000-0002-5653-8286}
%\authornotemark[*]
\authornote{Corresponding author: Yupeng Hu.}
\affiliation{%
\department{\normalsize School of Software}
  \institution{\normalsize Shandong University}
  \city{Jinan}
  \country{China}
  }
\email{huyupeng@sdu.edu.cn}

\author{Zixu Li}
\orcid{0009-0001-5136-159X}
\affiliation{
\department{\normalsize School of Software}
  \institution{\normalsize Shandong University}
  \city{Jinan}
  \country{China}
  }
\email{lizixu.cs@gmail.com}

\author{Zhiheng Fu}
\orcid{0009-0007-7724-5662}
\affiliation{
\department{\normalsize School of Software}
  \institution{\normalsize Shandong University}
  \city{Jinan}
  \country{China}
  }
\email{fuzhiheng8@gmail.com}

\author{Haokun Wen}
\orcid{0000-0003-0633-3722}
\affiliation{%
\department{\normalsize School of Computer Science and Technology}
  \institution{\normalsize Harbin Institute of Technology (Shenzhen)}
  \city{Shenzhen}
  \country{China}
  }
\affiliation{
\department{\normalsize School of Data Science}
  \institution{\normalsize City University of Hong Kong}
  \city{Hong Kong}
  \country{China}
  }
\email{whenhaokun@gmail.com}

\author{Weili Guan}
\orcid{0000-0002-5658-5509}
%\authornotemark[*]
% \authornotemark[1]
\affiliation{%
\department{\normalsize School of Information Science and Technology}
  \institution{\normalsize Harbin Institute of Technology (Shenzhen)}
  \city{Shenzhen}
  \country{China}
  }
\email{honeyguan@gmail.com}

\begin{abstract}
Composed Video Retrieval (CVR) is a challenging video retrieval task that utilizes multi-modal queries, consisting of a reference video and modification text, to retrieve the desired target video. The core of this task lies in understanding the multi-modal composed query and achieving accurate composed feature learning. Within multi-modal queries, the video modality typically carries richer semantic content compared to the textual modality. However, previous works have largely overlooked the disparity in information density between these two modalities. This limitation can lead to two critical issues: 1) \textbf{modification subject referring ambiguity} and 2) \textbf{limited detailed semantic focus}, both of which degrade the performance of CVR models. To address the aforementioned issues, we propose a novel CVR framework, namely the \underline{\textbf{H}}ierarchical \underline{\textbf{U}}ncertainty-aware \underline{\textbf{D}}isambiguation network (\textbf{HUD}). HUD is the first framework that leverages the disparity in information density between video and text to enhance multi-modal query understanding. It comprises three key components: (a) \textit{Holistic Pronoun Disambiguation}, (b) \textit{Atomistic Uncertainty Modeling}, and (c) \textit{Holistic-to-Atomistic Alignment}. By exploiting overlapping semantics through holistic cross-modal interaction and fine-grained semantic alignment via atomistic-level cross-modal interaction, HUD enables effective object disambiguation and enhances the focus on detailed semantics, thereby achieving precise composed feature learning. Moreover, our proposed HUD is also applicable to the Composed Image Retrieval (CIR) task and achieves state-of-the-art performance across three benchmark datasets for both CVR and CIR tasks.
The codes are available on~\href{https://zivchen-ty.github.io/HUD.github.io/}{https://zivchen-ty.github.io/HUD.github.io/}.
\end{abstract}


%%
%% Keywords. The author(s) should pick words that accurately describe
%% the work being presented. Separate the keywords with commas.

\maketitle

\section{Introduction}

\begin{figure}[ht!]
\vspace{-7pt}
	\includegraphics[width=0.95\linewidth]{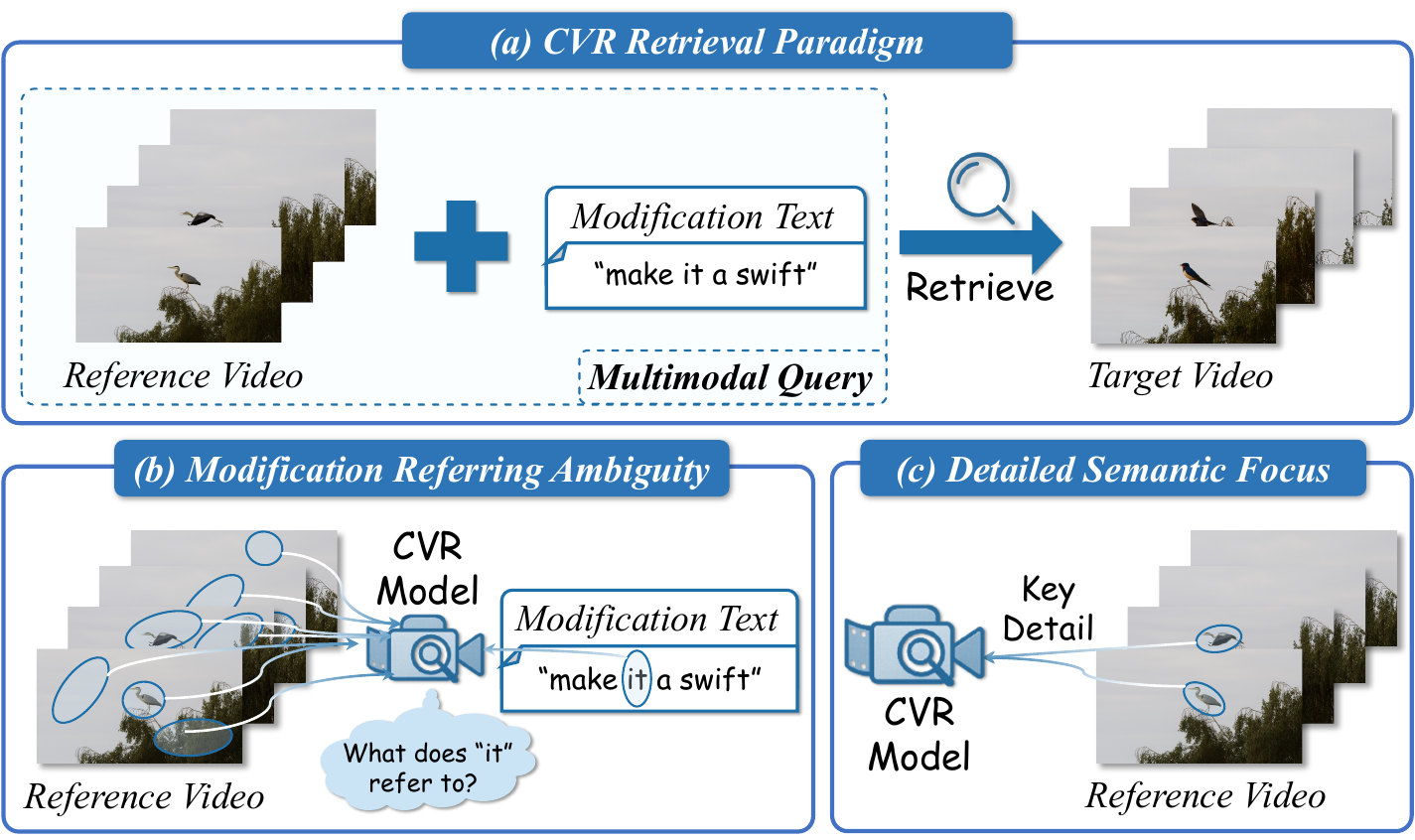}
    \vspace{-11pt}
	\caption{Illustrations of (a) an example of the CVR Retrieval Paradigm, the issue of (b) Modification Referring Ambiguity, and (c) Detailed Semantic Focus.}
        \vspace{-5pt}
	\label{fig:intro}
\end{figure}

In recent years, the explosive growth of video data has spurred widespread interest in video retrieval tasks among researchers, with the development of multimodal learning techniques~\cite{zeng2025FSDrive,jiang2025transforming,wang2025ascd,huang2025enhancing,zhang2024cf,sun2024dual,zeng2025janusvln,ni2025recondreamer,jing2023category,liu2025dp}. To address the diverse video retrieval requirements of users, \textit{Ventura et al.}~\cite{covr} proposed Composed Video Retrieval (CVR), which has garnered significant attention~\cite{covr-2,covr-cvpr,egocvr,fdca}. As shown in Figure~\ref{fig:intro}(a), unlike traditional unimodal video retrieval paradigms, CVR employs a multi-modal query comprised of a reference video and modification text. The modification text serves to express the user's desired modifications to the reference video. As a fundamental task in multi-modal interaction, research on CVR is expected to contribute to various downstream tasks and applications, such as video understanding~\cite{zhang2023multi,chen2025does,sun2023hierarchical,jing2023multimodal,kong2025modality,kong2025tuna}, information system~\cite{pu2025robust,edstedt2024dedode, huang2023robust,li2025frequency,zheng2025decoupled,ni2023feature,yi2025score,sha2022multi,zhang2024yoloppa,xu2025finmultitime,he2021semantic}, computer vision~\cite{zhao2024balf, sunmola2025surgical, yan2025hemora,zheng2024odtrack,zheng2022leveraging, huang2024ar,zhao2025drivedreamer4d,ni2025wonderturbo,ni2025wonderfree,huang_ai-augmented_2025,yang2024wcdt,wang2025mdanet, xin2025resurrect,awal2024vismin,wang2025target,wu2023towards,202508.0462}, and multimodal processing~\cite{pushe,zheng2025towards,zheng2023toward,wang2025unitmge,Zhang_2024_CVPR,cui2024correlation,Yin_2025_CVPR, wu2025evaluation}.

The key to the CVR task lies in understanding both the reference video and the modification text within the multi-modal query. Some pioneering works~\cite{covr-2,covr-cvpr,egocvr,fdca} have focused on enhancing the CVR models' understanding capability of multi-modal queries. Among them, \textit{Ventura et al.}~\cite{covr-2} directly applied BLIP~\cite{blip} and BLIP-2~\cite{blip-2} to CVR, leveraging query scoring~\cite{query_socre} to utilize information from multiple video frames. \textit{Thawakar et al.}~\cite{covr-cvpr} generated detailed language descriptions of the visual content to exploit the complementary contextual information inherent in videos. Moreover, other works~\cite{egocvr,fdca} have introduced new datasets that emphasize temporal understanding, fine-grained modifications, and event localization, thus further strengthening the models' multi-modal comprehension capabilities.
However, these methods have largely overlooked the disparity in information density between the video and text modalities. As the saying goes, ``a picture is worth a thousand words''. Compared to the text modality, the video modality often contains much richer semantics. This disparity further leads to two issues that adversely affect the performance of CVR models.

\textbf{Q1: Modification Subject Referring Ambiguity.}
As shown in Figure~\ref{fig:intro}(b), the referent of the pronoun ``it'' in the modification text is not explicitly stated. It is necessary to consider the context of the multi-modal query to infer that the pronoun ``it'' in the modification text refers to the ``bird'' in the video. Unfortunately, this is not an isolated case. 
As shown in Table~\ref{tab:ratio}, in the widely used CVR datasets (WebVid-CoVR~\cite{covr}), the proportion of modification texts that contain ambiguous pronouns (\textit{i.e.,} ``it'', ``them'') exceeds $34\%$ in the test set, respectively.
This complicates the model's capability of identifying visual modification objects and impedes its comprehension of multi-modal queries.

\begin{table}[htbp]
  \centering
        \vspace{-10pt}
  \caption{Ratios of modification texts containing ambiguous pronouns on WebVid-CoVR datasets.}
      \vspace{-10pt}
              \resizebox{0.87\linewidth}{!}{
    \begin{tabular}{c|ccc}
    \Xhline{1pt}
    \multirow{2}{*}{Dataset} & \multicolumn{3}{c}{WebVid-CoVR}  \\
\cline{2-4}          & Train & Val   & Test  \\
    \hline
    \#Ambiguous Pronouns Ratios& 44.76\% & 16.20\% & 34.66\%  \\
    \Xhline{1pt}
    \end{tabular}%
    }
     \vspace{-10pt}
  \label{tab:ratio}%
\end{table}%

\textbf{Q2: Limited Detailed Semantic Focus.}
As shown in Figure~\ref{fig:intro}(c),  the majority of the regions in the reference video are occupied by ``trees'' and ``sky''. However, the detail that requires modification is the ``bird'', which occupies only a small portion of the visual frame. For objects that occupy a minor visual proportion but play a crucial role in the modification request, we refer to them as ``key details''. Unfortunately, existing CVR models tend to allocate equal attention to all visual regions, which may result in insufficient focus on these key details and ultimately lead to bias in the composed feature.

To address the aforementioned issues, we propose a \underline{\textbf{H}}ierarchical \underline{\textbf{U}}ncertainty-aware \underline{\textbf{D}}isambiguation network (\textbf{HUD}), which is designed to improve multi-modal query understanding by leveraging the disparity in information density between video and text. It consists of three modules.
\textit{(a) Holistic Pronoun Disambiguation}, which aims to mine overlapping semantics across modalities from a holistic perspective, thereby indirectly disambiguating the referents of the pronouns in the modification text. This process mitigates the impact of modification subject referring ambiguity on multi-modal query understanding.
\textit{(b) Atomistic Uncertainty Modeling}, which leverages cross-modal interactions at the atomistic perspective to discern key detail semantics via uncertainty modeling. This serves as a complement to global semantics, enhancing the detail-focused capability of CVR models and addressing the issue of limited detailed semantic focus.
\textit{(c) Holistic-to-Atomistic Alignment}, which adaptively aligns the composed query representation with the target image by incorporating a learnable similarity bias between the holistic and atomistic levels.

The contributions of this work are summarized in three aspects:
 \begin{itemize}[leftmargin=8pt]
\item We propose a novel Composed Video Retrieval model, HUD, which, to the best of our knowledge, is the first CVR framework that enhances multi-modal query understanding by leveraging the disparity in information density between videos and texts.
\item The proposed HUD achieves modification object disambiguation through overlapping semantics in holistic-level interactions, and enhances the focus on detailed semantics by leveraging uncertainty semantics from atomistic-level interactions, thereby enabling precise composed learning.
\item Extensive qualitative and quantitative experiments are conducted on three widely used benchmark datasets, covering both CVR and CIR tasks. The results demonstrate the superiority of HUD.
\end{itemize}

\section{Related Work}

% \noindent
\textit{\textbf{Composed Video Retrieval.}}
% \subsection{Composed Video Retrieval.}
Similar to the composed image retrieval task~\cite{encoder,finecir,median,pair,tgcir,offset,lin2024fine}, the CVR task aims to design a model to interpret both the multi-modal video retrieval requirements and the modification text provided by the user using the reference video as context, and subsequently, it retrieves the corresponding target video from the database. 
\textit{Ventura et al.}\cite{covr, covr-2} first proposed this task and leveraged the generalization capabilities of visual language pre-trained backbones such as BLIP and BLIP-2 to comprehend multi-modal queries, adapting the backbone to the CVR task via a simple linear composition module. \textit{Thawakar et al.}\cite{covr-cvpr} utilized detailed captions to enrich the semantic expression of multi-modal queries, thereby enhancing CVR retrieval performance. Later, \textit{Hummel et al.}~\cite{egocvr} introduced the EgoCVR dataset, which extends the CVR task to ego-centric scenarios, with the aim of validating the model's ability to understand and compose fine-grained semantics from a ego-centric perspective.
Despite the significant progress these models have made in the development of the CVR task, they still overlook the issue of modification subject referring ambiguity and neglect specifically focus on fine-grained semantic details, resulting in limitations in CVR performance. In contrast, our HUD model disambiguates the text via probability embeddings and enhances focus on fine-grained semantics through a holistic-to-atomistic hierarchical alignment, thereby achieving superior retrieval performance.

\textit{\textbf{Probabilistic Representation for Uncertainty Modeling.}}
Our work focuses on mitigating the multi-modal query comprehension issues triggered by data uncertainty. This type of uncertainty generally originates from inherent data noise, such as annotation ambiguity, which has significantly impacted the field of deep learning technology~\cite{Chen_2025_CVPR,rong2025backdoor,huangexploring,huangnodes,huang2025final,liu2024dual,sun2023stepwise,zhong2025narrative,zhong2025enhancing}.
To address this uncertainty, previous studies employ probabilistic representations~\cite{wang2025humandreamer, wang2025humandreamer-x,wu2024novel,huang2024rock,liu2025fedmuon} to model uncertainty confidence.
For the visual domain, \textit{Oh et al.}\cite{HIB} first introduced a probabilistic embedding method to capture uncertainty relationships in image vision. Subsequently, \textit{He et al.}\cite{BoundingUncertainty} tackled the ambiguity of annotation boundaries in object detection data by learning transformation and localization variance of bounding boxes via KL loss, thereby effectively improving detection accuracy.
Moreover, \textit{Chun et al.}~\cite{pcme} extended probabilistic embeddings to the text-image retrieval task, further modeling the inherent ambiguities in both the visual and language modalities. Subsequent works further incorporate probabilistic embedding into cross-modal composition scenarios~\cite{mgur} to capture coarse-grained uncertainty in multi-modal composition.

Different from existing studies, our approach focuses on the more challenging CVR task, where videos participate in multi-modal query composition as a complex modality. It further exacerbates the uncertainty in the composed semantics. Meanwhile, existing methods typically only emphasize intra-modal interactions when modeling probabilistic embeddings, neglecting the important role of cross-modal semantic overlap in resolving ambiguity, thereby limiting further improvements in cross-modal retrieval performance.

\section{HUD}

\begin{figure*}[ht]
    \vspace{-10pt}
	\includegraphics[width=0.95\linewidth]{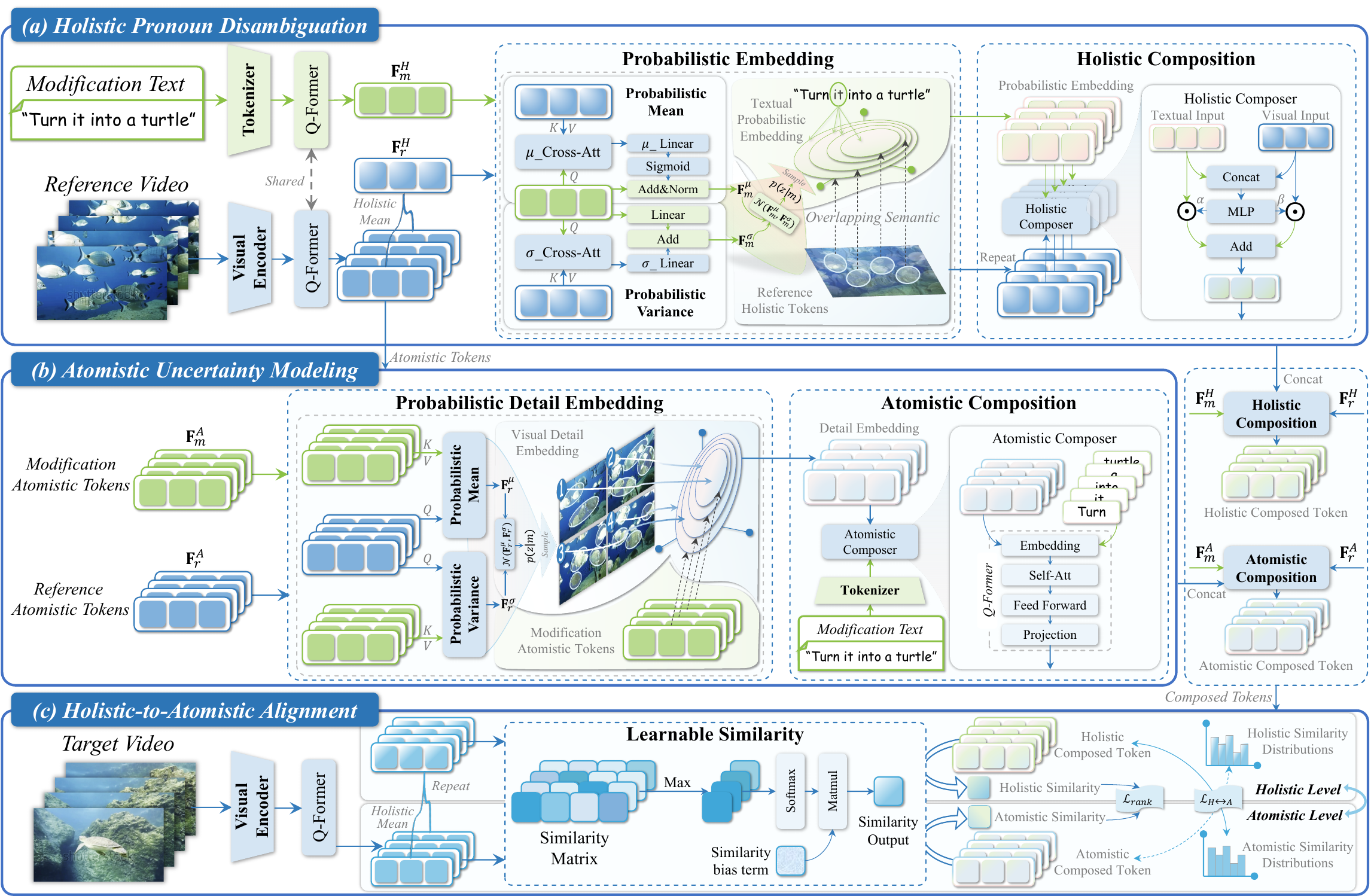}
    \vspace{-10pt}
	\caption{Overall framework of HUD consists of \textit{(a) Holistic Pronoun Disambiguation}, \textit{(b) Atomistic Uncertainty Modeling}, and \textit{(c) Holistic-to-Atomistic Alignment}.}
    \vspace{-15pt}
	\label{fig:Framework}
\end{figure*}

As the primary innovation, the proposed HUD model is designed to leverage the disparity in information density between video and text to improve multi-modal query comprehension. As depicted in Figure~\ref{fig:Framework}, HUD comprises three main modules:
\textit{(a) Holistic Pronoun Disambiguation}, which aims to extract overlapping semantics among modalities from a holistic perspective, thereby indirectly disambiguating the pronouns' referents in the modification text and mitigating the impact of modification subject referring ambiguity on multi-modal query understanding.
\textit{(b) Atomistic Uncertainty Modeling}, which aims to leverage cross-modal interactions at the atomistic level. It utilizes uncertainty modeling to discern key detail semantics and thereby serves as a supplement to global semantics. This module enhances the detail-focused capability of the CVR model and alleviates the problem of limited detailed semantic focus.
\textit{(c) Holistic-to-Atomistic Alignment}, which employs a learnable similarity bias to adaptively align the composed query representation with the target video at the holistic-to-atomistic level.
In this section, we first formulate the CVR task and subsequently elaborate on each module of the proposed HUD model.

\subsection{Problem Formulation}
The proposed HUD intends to address the CVR task, which aims to retrieve a target video that meets the requirements of a multi-modal query (including a reference video and a modification text). Suppose we are given a set of \(N\) triplets, denoted as \(\mathcal{T}=\{(x_{r}, x_{m}, x_{t})_n\}_{n=1}^{N}\), where \(x_{r}\), \(x_{m}\), and \(x_{t}\) represent the reference video, modification text, and target video, respectively. Essentially, our objective is to optimize a latent metric space in which the representation of the multi-modal query \((x_{r}, x_{m})\) is as close as possible to that of the corresponding target video \(x_{t}\), which can be formally expressed as, $ \mathcal{H}\left(x_r, x_m\right)\!\rightarrow\!\mathcal{H}\left(x_t\right),$
where \(\mathcal{H}\) denotes the embedding function that is optimized for both the multi-modal query and target video.

\subsection{\textbf{Holistic Pronoun Disambiguation}}
\label{sec:Holistic Pronoun Disambiguation}
In the CVR task, the modification text often contains ambiguous pronouns referring to modification objects. For example, in Figure~\ref{fig:intro}(b), the pronoun ``it'' appears in the modification text. It is difficult to determine which object in the reference video ``it'' refers to based solely on the modification text. Here, we propose to capture the overlapping semantics by cross-model interactions from a holistic perspective to take a detour to deal with this problem. Through such interactions between the modification text and the reference video, the model can infer that ``it'' refers to ``bird'' rather than ``tree'' or ``sky''.
Based on this intuition, we design the \textit{Holistic Pronoun Disambiguation} module. This module aims to alleviate the problem of modification subject referring ambiguity by exploiting cross-modal interactions to indirectly resolve pronoun ambiguities. The details of this module are described below.

\subsubsection{Holistic Token Extraction}
\label{subsec:HTE}
Initially, to facilitate cross-modal interactions, we first extract holistic visual and text features within a unified semantic space. Specifically, we employ the pre-trained Q-Former~\cite{blip-2} to extract tokens of the modification text and \(N_f\) frames sampled from the reference video, formulated as follows:
\begin{equation}
\textbf{F}_r^A=\operatorname{Q-Former}\left(\varPhi_{\mathbb{I}}\left(x_r\right)\right), \textbf{F}_m^A=\operatorname{Q-Former}\left(\varPhi_{\mathbb{T}}\left(x_m\right)\right),
% \left\{
% \begin{aligned}
%     & \\
%     & \\
% \end{aligned}
    \label{qformer}
% \right.
\end{equation}
where \(\textbf{F}_r^A \in \mathbb{R}^{(N_f \times L) \times D_A}\) and \(\textbf{F}_m^A \in \mathbb{R}^{L \times D_A}\) respectively denote the reference and modification atomistic tokens, \(N_f\) is the sampled frame number, \(L\) is learnable query number in the Q-Former, and \(D_A\) represents the atomistic embedding dimension. \(\varPhi_{\mathbb{I}}\) and \(\varPhi_{\mathbb{T}}\) denote the visual encoder and the text tokenizer, respectively. Then, to obtain the holistic representations of the video and text, we perform global average pooling and projection on the atomistic tokens to yield the reference and modification holistic tokens \(\textbf{F}_r^H, \textbf{F}_m^H \in \mathbb{R}^{D_H}\), where \(D_H\) represents the holistic embedding dimension.

\subsubsection{Probabilistic Embedding}
\label{subsec:Probabilistic Embedding}
Inspired by the application~\cite{pcme} of probabilistic distributions in image/video and text representations, we construct a probabilistic distribution for the modification text to enhance the understanding of ambiguous modification semantics. In addition, we utilize cross-attention to mine the overlapping semantics across modalities, which is capable of establishing semantic associations between text pronouns and their corresponding modification objects, and indirectly achieves pronoun disambiguation.

Specifically, we adopt the widely used Gaussian probability distribution~\cite{pcme,uatvr} to model the probability distribution of the modification text, denoted as \(p(z|m) \sim \mathcal{N}(\textbf{F}_m^\mu, \textbf{F}_m^\sigma)\), where \(\textbf{F}_m^\mu\) and \(\textbf{F}_m^\sigma\) denote the probabilistic mean and variance of the modification text, respectively, formulated as follows,
\begin{equation}
\left\{
\begin{aligned}
    &\textbf{F}_m^\mu=h^\mu_m\left(\operatorname{Cross-Att}_\mu\left(Q=\textbf{F}_m^H, KV=\textbf{F}_r^H\right)+\textbf{F}_m^H\right), \\
    &\textbf{F}_m^\sigma=h^\sigma_m\left(\operatorname{Cross-Att}_\sigma\left(Q=\textbf{F}_m^H, KV=\textbf{F}_r^H\right)+\textbf{F}_m^H\right), \\
\end{aligned}
    \label{mean_variance}
\right.
\end{equation}
where \(\textbf{F}_m^\mu, \textbf{F}_m^\sigma \in \mathbb{R}^{D_H}\), \(h^\mu_m\) is comprised of a linear layer, a sigmoid activation function, and LayerNorm, while \(h^\sigma_m\) is implemented as a single linear layer. So far, we have extracted the overlapping semantics between the modification text and the reference video through cross-attention and embedded these semantics into the Gaussian probability distribution of the modification text.

Subsequently, due to gradient propagation interruption caused by the sample randomness, we employ the reparametrization trick~\cite{reparametrization} to sample \textit{independent and identically distributed (i.i.d.) random variables} from the distribution \(p(z|m)\), formulated as,

\begin{equation}
\left\{
\begin{aligned}
    &\textbf{F}_m^{p}=\left[\textbf{F}_m^{p_1},...,\textbf{F}_m^{p_u},...,\textbf{F}_m^{p_U}\right]\overset{i.i.d.}{\sim}p(z|m), \\
    &\textbf{F}_m^{p_u}=\textbf{F}_m^\sigma \cdot \eta^u + \textbf{F}_m^\mu, \left(\eta^u\overset{i.i.d.}{\sim}\mathcal{N}\left(0,1\right)\right),\\
\end{aligned}
    \label{sample}
\right.
\end{equation}
where \(\textbf{F}_m^{p} \in \mathbb{R}^{U \times D_H}\) denotes the textual probabilistic embedding, with \(U\) representing the number of samples, and $\mathcal{N}(0,1)$ denotes the standard normal distribution.

\subsubsection{Holistic Composition}
\label{subsec:Holistic Composition}
To facilitate subsequent composed learning, we construct the holistic composed token from both the probability distribution and the original features. We first introduce the construction of the holistic composed token based on the probability distribution. 

Specifically, we leverage an MLP to learn the composition weight \(\textbf{w}^u\) between the \(u\)-th sampled textual probabilistic embedding \(\textbf{F}_m^{p_u}\) and the reference holistic token \(\textbf{F}_r^H\), formulated as,
\begin{equation}
\textbf{w}^u=\operatorname{MLP}\left(\left[\textbf{F}_r^H,\textbf{F}_m^{p_u}\right]\right),
    \label{weight_c}
\end{equation}
where \(\textbf{w}^u \in \mathbb{R}^{2 \cdot D_H}\). Subsequently, we perform a chunk operation on \(\textbf{w}^u\), denoting the resultant weights as \(\textbf{w}^u_\alpha, \textbf{w}^u_\beta \in \mathbb{R}^{D_H}\), and then aggregate these weights with \(\textbf{F}_m^{p_u}\) and \(\textbf{F}_r^H\) respectively to obtain the \(u\)-th holistic probabilistic composed token, formulated as,
\begin{equation}
\textbf{F}_c^{H_u}=\textbf{w}^u_\alpha\textbf{F}_m^{p_u}+\textbf{w}^u_\beta\textbf{F}_r^H,
    \label{holistic_compose}
\end{equation}
where $\textbf{F}_c^{H_u}\in \mathbb{R}^{D_H}$. 
Notably, in addition to the overlapping semantics, the reference video and the modification text also contain different semantics, which generally encapsulate crucial modification information (such as color changes and location modifications). While the overlapping semantics have been mined via the probability distribution, in order to extract the different semantics and ensure the semantic integrity of the multi-modal query, we apply an identical composition operation on the original features (\textit{i.e.,} \(\textbf{F}_r^H\) and \(\textbf{F}_m^H\), obtained in~\ref{subsec:HTE}), thereby yielding the holistic composed token $\textbf{F}_c^{H}=\left[\operatorname{Composition}\left(\textbf{F}_r^H, \textbf{F}_m^H\right), \textbf{F}_c^{H_1},...,\textbf{F}_c^{H_u},...,\textbf{F}_c^{H_U}\right]\in \mathbb{R}^{(1+U)\times D_H}$.

\subsection{\textbf{Atomistic Uncertainty Modeling}}
\label{sec:Atomistic Uncertainty Modeling}
As aforementioned, we note that certain objects in video frames may occupy a small area yet serve as important modification objects (for example, the ``bird'' in Figure~\ref{fig:intro}), which we refer to as ``key detail''. Existing CVR models typically allocate equal attention to all visual regions, potentially resulting in insufficient focus on these key detail objects and suboptimal composed learning. To enhance the model's focus on key details, we introduce the \textit{Atomistic Uncertainty Modeling} module. This module aims to leverage cross-modal interactions from an atomistic perspective to discern key detail semantics through uncertainty modeling, serving as a complement to holistic semantics. Thereby, it improves the detail-focused capability of HUD, effectively addressing the issue of limited detailed semantic focus. We describe this module in detail as follows.

\subsubsection{Probabilistic Detail Embedding}
Since the reference video encompasses many details, only the modification objects qualify as the ``key details''. Therefore, to amplify the model's focus on detail, it is necessary to discern the ``key detail'' among the myriad details. Specifically, we introduce an atomistic uncertainty modeling for visual detail embedding. Similar to the \textit{Probabilistic Embedding} (described in~\ref{subsec:Probabilistic Embedding}), we first construct a Gaussian probability distribution for the reference video, denoted as \( p(z|r) \sim \mathcal{N}(\textbf{F}^\mu_r, \textbf{F}_r^\sigma) \), where \(\textbf{F}^\mu_r\) and \(\textbf{F}_r^\sigma\) denote the probabilistic mean and variance of the reference video, respectively, formulated as,
\begin{equation}
\left\{
\begin{aligned}
    &\textbf{F}_r^\mu=h^\mu_r\left(\operatorname{Cross-Att}_\mu\left(Q=\textbf{F}_r^A, KV=\textbf{F}_m^A\right)+\textbf{F}_r^A\right), \\
    &\textbf{F}_r^\sigma=h^\sigma_r\left(\operatorname{Cross-Att}_\sigma\left(Q=\textbf{F}_r^A, KV=\textbf{F}_m^A\right)+\textbf{F}_r^A\right), \\
\end{aligned}
    \label{mean_variance_}
\right.
\end{equation}
where \(h^\mu_r\) and \(h^\sigma_r\) share the same architecture as \(h^\mu_m\) and \(h^\sigma_m\) in Eq.$($\ref{mean_variance}$)$. \(\textbf{F}_r^A\) and \(\textbf{F}_m^A\) denote the reference and modification atomistic tokens obtained in Eq.$($\ref{qformer}$)$, and \(\textbf{F}_r^\mu, \textbf{F}_r^\sigma \in \mathbb{R}^{(N_f \times L) \times D_A}\). 
Subsequently, as each reference atomistic token may contain crucial modification detail semantics, we sample \textit{i.i.d.} random variables for each token from the distribution \(p(z|r)\) in order to model the visual detail embedding for each token, formulated as follows,
\begin{equation}
\left\{
\begin{aligned}
    &\textbf{F}_r^{p}=\left[\textbf{F}_r^{p_1},...,\textbf{F}_r^{p_l},...,\textbf{F}_r^{p_{N_f\times L}}\right]\overset{i.i.d.}{\sim}p(z|r), \\
    &\textbf{F}_r^{p_l}=\textbf{F}_r^\sigma \cdot \eta^l + \textbf{F}_r^\mu, \left(\eta^l\overset{i.i.d.}{\sim}\mathcal{N}\left(0,1\right)\right),\\
\end{aligned}
    \label{sample_}
\right.
\end{equation}
where $\textbf{F}_r^{p}\in \mathbb{R}^{(N_f\times L)\times D_A}$ denotes the visual detail embedding.

\subsubsection{Atomistic Composition}
To ensure that the composed token captures both the holistic and atomistic semantics of the multi-modal query, we construct the atomistic composed token corresponding to the holistic composed token construction (detailed in~\ref{subsec:Holistic Composition}). 
In contrast to the holistic composition, the atomistic composition focuses more on the fine-grained modification details. To this end, we utilize the fine-grained attention mechanism of the \mbox{Q-Former}~\cite{blip-2} to establish detail-level associations between the visual and textual modalities, thereby enabling precise detail extraction and multi-modal composition.

Specifically, we use the visual detail embedding \(\textbf{F}_r^{p}\) as the query embedding and feed it together with the modification text token sequence \(\varPhi_{\mathbb{T}}(x_m)\) into the Q-Former for fine-grained interaction, formulated as follows,
\begin{equation}
\textbf{F}^p_c=\operatorname{Q-Former}\left(\textbf{F}_r^{p}, \varPhi_{\mathbb{T}}\left(x_m\right)\right),
    \label{atomistic_QFormer_}
\end{equation}
where \(\textbf{F}^p_c \in \mathbb{R}^{(N_f \times L) \times D_A}\) denotes the atomistic uncertainty composed token, and \(\varPhi_{\mathbb{T}}\) represents the text tokenizer. Similar to the \textit{Holistic Composition}, to retain the different semantics from the original atomistic tokens, we perform fine-grained interaction between the original reference atomistic tokens \(\textbf{F}_r^{A}\) and the modification text token sequence. Moreover, to facilitate the subsequent holistic-to-atomistic alignment, we align the feature dimensions of the atomistic and holistic tokens via projection~\cite{blip-2}, formulated as,
\begin{equation}
\textbf{F}^A_c=\operatorname{Proj}\left(\left[\operatorname{Q-Former}\left(\textbf{F}_r^{A}, \varPhi_{\mathbb{T}}\left(x_m\right)\right), \textbf{F}^p_c\right]\right),
    \label{atomistic_QFormer}
\end{equation}
where $\textbf{F}^A_c\in \mathbb{R}^{(N_f\times 2L)\times D_H}$ denotes the atomistic composed token.

\subsection{Holistic-to-Atomistic Alignment}
\label{sec:Holistic-to-Atomistic Alignment}
Due to the inherent uncertainty of the previously sampled \textit{i.i.d.} random variables, directly computing the feature similarity between the multi-modal query and the target video may introduce semantic bias. To address this issue, we design the \textit{Holistic-to-Atomistic Alignment} module, which utilizes a learnable bias term in the similarity computation. Thus, it eliminates any semantic bias imposed by the probabilistic distribution, and guides the holistic and atomistic composed tokens to adaptively align with the target video.

Specifically, we first adopt the same process as in Eq.$($\ref{qformer}$)$ to obtain the holistic and atomistic target tokens, and extend their channel dimensions to match those of the holistic and atomistic composed tokens, denoted as \(\textbf{F}^H_t \in \mathbb{R}^{(1+U) \times D_H}\) and \(\textbf{F}^A_t \in \mathbb{R}^{(N_f \times 2L) \times D_H}\), respectively. Subsequently, taking the holistic similarity computation as an example, we define the similarity bias calculation function as,
\begin{equation}
\mathcal{B}\left(\textbf{F}^H_c, \textbf{F}^H_t\right)=\operatorname{softmax}\left(\max^{(1+U)}_{i=1}\left(\textbf{F}^H_{c(i)}{\textbf{F}^{H^\top}_{t(i)}}\right)\right)+\textbf{b}_H,
    \label{holistic_score_}
\end{equation}
where \(\textbf{b}_H \in \mathbb{R}^{(1+U)}\) and \(\textbf{F}^H_{c(i)}, \textbf{F}^H_{t(i)}\) denote the \(i\)-th token of the holistic composed token and target token, respectively. Then, this similarity bias term is incorporated into the original similarity computation as a weighting factor for holistic similarity. Consequently, we define the learnable similarity function as follows,
\begin{equation}
\mathcal{S}\left(\textbf{F}^H_c, \textbf{F}^H_t\right)=\operatorname{sum}\left(\mathcal{B}\left(\textbf{F}^H_c, \textbf{F}^H_t\right) \cdot \max^{(1+U)}_{i=1}\left(\textbf{F}^H_{c(i)}{\textbf{F}^{H^\top}_{t(i)}}\right)\right) \in \mathbb{R}^{1}.
    \label{holistic_score}
\end{equation}

In the same manner, we can obtain the atomistic similarity, denoted as \(\mathcal{S}\left(\textbf{F}^A_c, \textbf{F}^A_t\right) \in \mathbb{R}^{1}\). Subsequently, to simultaneously promote both the holistic and atomistic composed tokens closer to the corresponding target video, we employ a batch-based classification loss~\cite{batch-based-classification-loss} using both holistic and atomistic similarity, formulated as, 
\begin{equation}
% \fontsize{9pt}{9pt}
\small
    \mathcal{L}_{rank} \!=\! \frac{1}{B} \!\sum_{i=1}^{B} \!-\!\log\! \left\{ \frac{\exp \left\{ \left( \mathcal{S}\left(\textbf{F}^H_{ci}, \textbf{F}^H_{ti}\right)+\mathcal{S}\left(\textbf{F}^A_{ci}, \textbf{F}^A_{ti}\right)\right)  / \tau\right\}}{ \sum_{j=1}^{B} \exp \left\{ \left( \mathcal{S}\left(\textbf{F}^H_{ci}, \textbf{F}^H_{tj}\right)+\mathcal{S}\left(\textbf{F}^A_{ci}, \textbf{F}^A_{tj}\right)\right) / \tau \right\}  } \right\},
    \label{bbc}
\end{equation}
where \(B\) denotes the batch size and \(\tau\) represents the temperature coefficient. \(\textbf{F}^H_{ci}\) and \(\textbf{F}^H_{ti}\) denote the \(i\)-th holistic composed token and target token in the batch, respectively.

Additionally, since the holistic and atomistic tokens are modeled separately, their semantic distributions may be inconsistent. To ensure that the atomistic semantics captured at the fine-grained level do not conflict with the holistic semantics, we introduce a distribution regularization, thereby promoting distributional consistency across the semantic levels. 
Specifically, let \(\textbf{s}^H_{ci} = \left[s^{H1}_{ci}, \ldots, s^{Hj}_{ci},\ldots, s^{HB}_{ci}\right]\) denote the similarity degree distribution of the \(i\)-th holistic composed token in the batch, where \(s^{Hj}_{ci}\) represents the similarity degree between the \(i\)-th holistic composed token and the \(j\)-th holistic target token, formulated as,
\begin{equation}
\small
s^{Hj}_{ci} =  \frac{\exp \left\{ \mathcal{S}\left(\textbf{F}^H_{ci}, \textbf{F}^H_{tj}\right)  / \tau\right\}}{ \sum_{b=1}^{B} \exp \left\{ \mathcal{S}\left(\textbf{F}^H_{ci}, \textbf{F}^H_{tb}\right) / \tau \right\}  }. 
\label{sim_degree}
\end{equation}

Similarly, we obtain the similarity degree distribution of the \(i\)-th holistic target token in the batch, denoted as \(\textbf{s}^H_{ti} = \left[s^{H1}_{ti}, \ldots, s^{HB}_{ti}\right]\), as well as the similarity degree distributions for the atomistic composed and target tokens in the batch, denoted as \(\textbf{s}^A_{ci}\) and \(\textbf{s}^A_{ti}\), respectively. Then, we define the distribution regularization as follows,
\begin{equation}
\mathcal{L}_{H\leftrightarrow A}= \frac{1}{2B}\sum_{i=1}^{B}\left(\operatorname{KL}\left(\textbf{s}^A_{ti} \| \textbf{s}^H_{ci}\right) + \operatorname{KL}\left(\textbf{s}^H_{ti} \| \textbf{s}^A_{ci}\right) \right).
\label{kl}
\end{equation}

Finally, we obtain the optimization function for HUD as follows,
\begin{equation}
\mathbf{\Theta^{*}}=
\underset{\mathbf{\Theta}}{\arg \min } \left( \mathcal{L}_{rank} + \kappa \mathcal{L}_{H\leftrightarrow A}\right),
\label{optimization}
\end{equation}
where $\mathbf{\Theta}$ is the HUD parameter to be learned and $\kappa$ is the trade-off hyper-parameter.

\section{Experiment}
In this section, we first outline the experimental settings, followed by a detailed presentation of the experimental results and corresponding analyses.

\subsection{Experimental Settings}
\subsubsection{Datasets.}
The proposed HUD framework is not only suitable for the dedicated CVR task but also adaptable to the CIR task. Following previous works~\cite{covr-cvpr,covr, covr-2}, we select the large-scale WebVid-CoVR~\cite{covr} for evaluation on the CVR task. For the CIR task, we select the widely used fashion-domain dataset FashionIQ~\cite{FashionIQ} and open-domain dataset CIRR~\cite{cirr} for evaluation.

\subsubsection{Implementation Details.}
Following previous works~\cite{covr-2}, we utilize the BLIP-2~\cite{blip-2} finetuned on the COCO dataset at $364$ pixels as the backbone of HUD and freeze the ViT during training for computational efficiency. The frame number $N_f$ is set to $4$ on CVR datasets. The learnable query number $L$ of Q-Former is $32$. The sample number $U$ of the textual probabilistic embedding is set to $6$ on the fashion-domain dataset FashionIQ, and $8$ on open-domain datasets, WebVid-CoVR, and CIRR. 
The trade-off hyper-parameter $\kappa$ is searched using a grid search, and finally confirmed as $\kappa=0.5$. And the temperature coefficient $\tau$ is set to $0.1$. We train HUD with a batch size $B=64$ and utilize the AdamW optimizer with the learning rate of $1e-5$. 
All experiments were performed on an NVIDIA A40 GPU with $48$GB memory, and trained for $5$ and $10$ epochs on CVR and CIR tasks, respectively.

\subsubsection{Evaluation.} 
To ensure a fair comparison among the evaluated models, we adhered to the standard evaluation protocols for each dataset and report Recall@\(k\) (abbreviated as R@\(k\)). For the WebVid-CoVR dataset, we report the R@$k$ with $k = 1,5,10,50$ along with their mean values.
In the case of FashionIQ, R@$10$ and R@$50$ are reported on each category, as well as their corresponding average values for each category. For the CIRR dataset, we include R@\(k\) for \(k = 1, 5, 10, 50\), R\(_{subset}\)@\(k\) for \(k = 1, 2, 3\), and the mean of R@$5$ and R\(_{subset}\)@$1$.

% Table generated by Excel2LaTeX from sheet 'FashionIQ'
\begin{table}[htbp]
  \centering
  \caption{Performance comparison on the test set of the CVR dataset, WebVid-CoVR. The overall best results are in bold, while the best results over baselines are underlined.}
  \vspace{-10pt}
    \resizebox{0.95\linewidth}{!}{
    \begin{tabular}{l|cccc|c}
    \Xhline{1pt}
    \multicolumn{1}{c|}{\multirow{3}{*}{Method}} & \multicolumn{5}{c}{WebVid-CoVR-Test} \\
\cline{2-6}          & \multicolumn{4}{c|}{R@$k$}      & \multirow{2}{*}{Avg.}\\
\cline{2-5}     & $k$=$1$   & $k$=$5$   & $k$=$10$  & $k$=$50$  &      \\
    \hline
    \hline
   \rowcolor[rgb]{ .949,  .949,  .949} \multicolumn{6}{c}{\textit{Pre-trianed Models}}\\
    CLIP~\cite{clip}\small{\textcolor{gray}{(ICML'21)}}  & 44.37 & 69.13 & 77.62 & 93.00 & 71.03 \\
    BLIP~\cite{blip}\small{\textcolor{gray}{(ICML'22)}}  & 45.46 & 70.46 & 79.54 & 93.27 & 72.18  \\
    % EgoVLPv2~\cite{egovlpv2}\small{\textcolor{gray}{(CVPR'23)}} & -     & -     & -     & -     & -     \\
    % LanguageBind~\cite{languagebind}\small{\textcolor{gray}{(ICLR'24)}} & -     & -     & -     & -     & -     \\
     \hdashline
    \rowcolor[rgb]{ .949,  .949,  .949} \multicolumn{6}{c}{\textit{CVR Models}}\\
    CoVR~\cite{covr}\small{\textcolor{gray}{(AAAI'24)}} & 53.13 & 79.93 & 86.85 & 97.69 & 79.40 \\
    CoVR\_Enrich~\cite{covr-cvpr}\small{\textcolor{gray}{(CVPR'24)}} &\underline{60.12} & \underline{84.32} & 91.27 & \underline{98.72} & \underline{83.61}  \\
    CoVR-2~\cite{covr-2}\small{\textcolor{gray}{(TPAMI'24)}} & 59.82 & 83.84 & \underline{91.28} & 98.24 & 83.30 \\
    FDCA~\cite{fdca}\small{\textcolor{gray}{(ICLR'25)}} & 54.80 & 82.27 & 89.84 & 97.70 & 81.15 \\
    \hline
    \hline
        \rowcolor[rgb]{ .851,  .851,  .851}
    \textbf{HUD (Ours)} & \textbf{63.38} & \textbf{86.93} & \textbf{92.29} & \textbf{98.76} & \textbf{85.34} \\
    \Xhline{1pt}
    \end{tabular}%
    }
     \vspace{-10pt}
  \label{tab:cvr}%
\end{table}%

% Table generated by Excel2LaTeX from sheet 'FashionIQ'
\begin{table*}[htbp]
  \centering
  \caption{Performance comparison on the CIR dataset, FashionIQ and CIRR, relative to R@$k$($\%$). The overall best results are in bold, while the best results over baselines are underlined.}
    \vspace{-10pt}
        \resizebox{0.96\textwidth}{!}{
    \begin{tabular}{l|cc|cc|cc|cc|cccc|ccc|c}
    \Xhline{1pt}
    \multicolumn{1}{c|}{\multirow{3}{*}{Method}} & \multicolumn{8}{c|}{FashionIQ}                                & \multicolumn{8}{c}{CIRR} \\
\cline{2-17}          & \multicolumn{2}{c|}{Dresses} & \multicolumn{2}{c|}{Shirts} & \multicolumn{2}{c|}{Tops\&Tees} & \multicolumn{2}{c|}{Avg.} & \multicolumn{4}{c|}{R@$k$}      & \multicolumn{3}{c|}{R$_{subset}$@$k$} & \multirow{2}{*}{Avg.} \\
\cline{2-16}          & R@$10$  & R@$50$  & R@$10$  & R@$50$  & R@$10$  & R@$50$  & R@$10$ & R@$50$  & $k$=$1$   & $k$=$5$   & $k$=$10$  & $k$=$50$  & $k$=$1$   & $k$=$2$   & $k$=$3$   &  \\
    \hline
    \hline
\rowcolor[rgb]{ .949,  .949,  .949} \multicolumn{17}{c}{\textit{CIR Models}}\\
    TG-CIR~\cite{tgcir}\small{\textcolor{gray}{(ACM MM'23)}} & 45.22  & 69.66  & 52.60  & 72.52  & 56.14  & 77.10  & 51.32  & 73.09  & 45.25  & 78.29  & 87.16  & 97.30  & 72.84  & 89.25  & 95.13  & 75.57  \\
    MGUR~\cite{mgur}\small{\textcolor{gray}{(ICLR'24)}} & 32.61  & 61.34  & 33.23  & 62.55  & 41.40  & 72.51  & 35.75  & 65.47  & -     & -     & -     & -     & -     & -     & -     & - \\
    % SSN~\cite{ssn}\small{\textcolor{gray}{(AAAI'24)}} & 34.36  & 60.78  & 38.13  & 61.83  & 44.26  & 69.05  & 38.92  & 63.89  & 43.91  & 77.25  & 86.48  & 97.45  & 71.76  & 88.63  & 95.54  & 74.51  \\
    SADN~\cite{sadn}\small{\textcolor{gray}{(ACM MM'24)}} & 40.01  & 65.10  & 43.67  & 66.05  & 48.04  & 70.93  & 43.91  & 67.36  & 44.27  & 78.10  & 87.71 & 97.89 & 72.34 & 88.70  & 95.23 & 75.22  \\
    SPRC~\cite{sprc}\small{\textcolor{gray}{(ICLR'24)}} & 49.18  & 72.43  & 55.64  & 73.89  & 59.35  & 78.58  & 54.72  & 74.97  & \underline{51.96} & \underline{82.12} & \underline{89.74} & 97.69 & \underline{80.65} & \underline{92.31} & \underline{96.60} & \underline{81.39}  \\
    LIMN~\cite{limn}\small{\textcolor{gray}{(TPAMI'24)}} & 50.72  & 74.52  & 56.08  & 77.09  & 60.94  & 81.85  & 55.91  & 77.82  & 43.64  & 75.37  & 85.42  & 97.04  & 69.01  & 86.22  & 94.19  & 72.19  \\
    LIMN+~\cite{limn}\small{\textcolor{gray}{(TPAMI'24)}} & \underline{52.11}  & \underline{75.21}  & \underline{57.51}  & \underline{77.92}  & \underline{62.67}  & \underline{82.66}  & \underline{57.43}  & \underline{78.60}  & 43.33  & 75.41  & 85.81  & 97.21  & 69.28  & 86.43  & 94.26  & 72.35  \\
    IUDC~\cite{iudc}\small{\textcolor{gray}{(TOIS'25)}} &35.22& 61.90 &41.86& 63.52& 42.19& 69.23& 39.76& 64.88 & -     & -     & -     & -     & -     & -     & -     & -\\
    ENCODER~\cite{encoder}\small{\textcolor{gray}{(AAAI'25)}} & 51.51  & 76.95  & 54.86  & 74.93  & 62.01  & 80.88  & 56.13  & 77.59  & 46.10  & 77.98  & 87.16  & 97.64  & 76.92  & 90.41  & 95.95  & 77.45  \\
    \hdashline
\rowcolor[rgb]{ .949,  .949,  .949} \multicolumn{17}{c}{\textit{CVR Models}}\\
    CoVR~\cite{covr}\small{\textcolor{gray}{(AAAI'24)}} & 44.55  & 69.03  & 48.43  & 67.42  & 52.60  & 74.31  & 48.53  & 70.25  & 49.69  & 78.60  & 86.77  & 94.31  & 75.01  & 88.12  & 93.16  & 74.51  \\
    CoVR\_Enrich~\cite{covr-cvpr}\small{\textcolor{gray}{(CVPR'24)}} & 46.12  & 69.52  & 49.61  & 68.88  & 53.79  & 74.74  & 49.84  & 71.05  & 51.03  & -     & 88.93  & 97.53  & 76.51  & -     & 95.76  & -  \\
    CoVR-2~\cite{covr-2}\small{\textcolor{gray}{(TPAMI'24)}} & 46.53  & 69.60  & 51.23  & 70.64  & 52.14  & 73.27  & 49.97  & 71.17  & 50.43 & 81.08 & 88.89 & \underline{98.05} & 76.75 & 90.34 & 95.78 & 78.92 \\
    \rowcolor[rgb]{ .851,  .851,  .851}
    \textbf{HUD (Ours)} & \textbf{54.34} & \textbf{76.85} & \textbf{62.07} & \textbf{80.52} & \textbf{65.12} & \textbf{84.40} & \textbf{60.51} & \textbf{80.59} & \textbf{52.84} & \textbf{82.15} & \textbf{90.00} & \textbf{98.07} & \textbf{81.64} & \textbf{93.35} & \textbf{97.64} & \textbf{81.89} \\
    \Xhline{1pt}        
    \end{tabular}%
    }
      \vspace{-5pt}
  \label{tab:cir}%
\end{table*}%

\subsection{Performance Comparison}
To validate the superior performance of HUD, we conduct extensive performance comparisons with other models on CVR and CIR tasks.

\subsubsection{\textbf{On CVR Task.}} As illustrated in Table~\ref{tab:cvr}, we select two categories of models to compare with our proposed HUD: pre-trained models (\textit{e.g.}, BLIP~\cite{blip} and CLIP~\cite{clip}) and several CVR models (\textit{e.g.}, CoVR~\cite{covr}, and FDCA~\cite{fdca}). From the analysis of the comparative results in the table, we can obtain the following observations.
\textbf{1)} HUD achieves the best performance across all metrics on both CVR datasets. Specifically, on the WebVid-CoVR, HUD attains improvements of $2.07$\% on the average metric, respectively. This demonstrates that leveraging the disparity in information density between video and text enables HUD to effectively understand multi-modal query semantics.
\textbf{2)} CoVR\_Enrich attains sub-optimal performance on WebVid-CoVR, which might be attributed to its reliance on additional generated video captions to enhance the model's perception of visual contextual information. In contrast, our proposed HUD requires no extra inputs and achieves superior performance solely through the designed holistic pronoun disambiguation and atomistic uncertainty modeling modules.

\subsubsection{\textbf{On CIR Task.}} 
In addition to the CVR task, we also conduct comprehensive experiments on the CIR task to demonstrate HUD's generalization capability for CIR. As illustrated in Table~\ref{tab:cir}, we select both CIR models (presented in the upper part of the table, \textit{e.g.,} ENCODER~\cite{encoder}, SPRC~\cite{sprc}) and CVR models (shown in the lower part of the table, \textit{e.g.,} CoVR~\cite{covr}, CoVR-2~\cite{covr-2}). From the results in the table, we derive the following important analytical insights.
\textbf{1)} HUD achieves competitive performance across all metrics on both CIR datasets. Specifically, compared with the sub-optimal methods, HUD attains a relative improvement of $5.36$\% in Avg.R@$10$ on FashionIQ and an $1.69$\% relative improvement in R@$1$ on CIRR. This indicates that the holistic-to-atomistic probabilistic modeling of HUD exhibits excellent domain generalizability.
\textbf{2)} Previous CVR models typically underperform on CIR tasks when compared to dedicated CIR models. This may be due to that they only process visual data from a global perspective and rely on the recurring presence of key objects across multiple video frames. Thus, they neglect attention to intra-frame visual details. In contrast, HUD simultaneously focuses on both holistic and atomistic details, thereby catering to the needs of both CVR and CIR tasks. This demonstrates that HUD has a strong generalization capability for semantic understanding across different visual modalities.

% Table generated by Excel2LaTeX from sheet 'Albation'
\begin{table}[htbp]
  \centering
  \caption{Ablation study on the CVR datasets, WebVid-CoVR, and CIR datasets, FashionIQ and CIRR. $\Delta$ denotes the performance drop of the compared derivatives and is marked with \textcolor{defgreen4}{\textit{the green background}}.}
    \vspace{-10pt}
            \resizebox{0.92\linewidth}{!}{
    \begin{tabular}{cc|cc|cc|cc}
    \Xhline{1pt}
    \multicolumn{1}{c|}{\multirow{2}{*}{D\#}} & \multirow{2}{*}{Derivatives} & \multicolumn{2}{c|}{WebVid-CoVR} & \multicolumn{2}{c|}{FashionIQ} & \multicolumn{2}{c}{CIRR} \\
\cline{3-8}    \multicolumn{1}{c|}{} &       & Avg.  & $\Delta$ & Avg. & $\Delta$ & Avg. & $\Delta$ \\
    \hline
    \hline
    \rowcolor[rgb]{ .949,  .949,  .949} \multicolumn{8}{c}{\textit{\textbf{M1: }Ablation on Holistic Pronoun Disambiguation}} \\
    \multicolumn{1}{c|}{(1)} & \multicolumn{1}{l|}{w/o\_H\_ Prob } & 83.49  & \cellcolor{defgreen2}-1.85  & 67.81  & \cellcolor{defgreen2}-2.74  &   78.45    &\cellcolor{defgreen3} -3.44  \\
    \multicolumn{1}{c|}{(2)} & \multicolumn{1}{l|}{w/o\_H\_Compose} & 84.07  & \cellcolor{defgreen1}-1.27  & 67.78  & \cellcolor{defgreen3}-2.77  & 79.04  &\cellcolor{defgreen2} -2.85  \\
    \multicolumn{1}{c|}{(3)} & \multicolumn{1}{l|}{w/o\_Holistic\_Level} & 82.98  & \cellcolor{defgreen3}-2.36  & 66.06  & \cellcolor{defgreen4}-4.49  &  77.56     &\cellcolor{defgreen4} -4.33   \\
    \rowcolor[rgb]{ .949,  .949,  .949} \multicolumn{8}{c}{\textit{\textbf{M2: }Ablation on Atomistic Uncertainty Modeling}} \\
    \multicolumn{1}{c|}{(4)} & \multicolumn{1}{l|}{w/o\_A\_ Detail  } & 83.88  & \cellcolor{defgreen1}-1.46  & 68.96  & \cellcolor{defgreen2}-1.59  &  79.64     & \cellcolor{defgreen2} -2.25  \\
    \multicolumn{1}{c|}{(5)} & \multicolumn{1}{l|}{w/o\_A\_Compose} & 83.75  & \cellcolor{defgreen2}-1.59  & 65.42  & \cellcolor{defgreen3}-5.13  &   78.93   & \cellcolor{defgreen3} -2.96 \\
    \multicolumn{1}{c|}{(6)} & \multicolumn{1}{l|}{w/o\_Atomistic\_Level} & 82.17  & \cellcolor{defgreen3}-3.17  & 64.29  & \cellcolor{defgreen4}-6.26  &  77.98     & \cellcolor{defgreen4}-3.91  \\
    \rowcolor[rgb]{ .949,  .949,  .949} \multicolumn{8}{c}{\textit{\textbf{M3: }Ablation on Holistic-to-Atomistic Alignment}} \\
    \multicolumn{1}{c|}{(7)} & \multicolumn{1}{l|}{w/o\_Bias} & 83.51  & \cellcolor{defgreen2}-1.83  & 68.32  & \cellcolor{defgreen2}-2.23  & 79.63  & \cellcolor{defgreen2}-2.26 \\
    \multicolumn{1}{c|}{(8)} & \multicolumn{1}{l|}{w/o\_$\mathcal{L}_{H\leftrightarrow A}$} & 84.04  & \cellcolor{defgreen1}-1.30  & 69.41  & \cellcolor{defgreen1}-1.14  & 79.87  & \cellcolor{defgreen1}-2.02\\
    \multicolumn{1}{c|}{(9)} & \multicolumn{1}{l|}{w/o\_$\mathcal{L}_{rank}$} & 81.14  & \cellcolor{defgreen3}-4.20  & 65.10  & \cellcolor{defgreen4}-5.45  & 72.77  & \cellcolor{defgreen4}-9.12\\
    \rowcolor[rgb]{ .851,  .851,  .851}
    \multicolumn{2}{c|}{\textbf{HUD}} & \textbf{85.34} & \cellcolor{defyellow}0.00  & \textbf{70.55 } & \cellcolor{defyellow}0.00  & \textbf{81.89 } & \cellcolor{defyellow}0.00  \\
    \Xhline{1pt}
    \end{tabular}%
    }
    \vspace{-12pt}
  \label{tab:ablation}%
\end{table}%

\subsection{Ablation Study}
To validate the effectiveness of each module designed in our proposed HUD, we conduct comprehensive ablation experiments on CVR datasets by comparing HUD with its derivatives, which can be divided into three groups according to the modules as follows:

\noindent
\textit{\textbf{$\bullet$ M1:Ablation on Holistic Pronoun Disambiguation.}} This group focuses on the ablation on the holistic level of HUD and includes the following derivatives.
\textbf{D\#(1) w/o\_H\_Prob} removes the textual probabilistic embedding to confirm its effect. \textbf{D\#(2) w/o\_H\_Compose} aims to validate the effect of holistic composition via replacing it with simple addition. \textbf{D\#(3) w/o\_Holistic\_Level} is conducted to explore the whole holistic impact via removing the \textit{Holistic Pronoun Disambiguation} module and the corresponding holistic features.

From the results of \textit{\textbf{M1}} in Table~\ref{tab:ablation}, we can obtain the following observations.
\textbf{1)} Both \textbf{D\#(1)} and \textbf{D\#(2)} show significant performance degradation compared to the complete HUD model, which indicates that the probabilistic embedding incorporated in HUD indeed enhances the understanding of ambiguous semantics and that the holistic composition facilitates the learning of more accurate modification semantics in the composed features. Moreover, on the CIR datasets, both \textbf{D\#(1)} and \textbf{D\#(2)} exhibit the most significant drop observed on the open-domain CIR dataset, CIRR. This is reasonable due to that in the open-domain CIR dataset, the object described in the modification text might correspond to multiple entities in the image, leading to ambiguity in determining the modification object. The incorporation of the textual probabilistic embedding along with the holistic composition effectively resolves this ambiguity, resulting in performance enhancement.
\textbf{2)} \textbf{D\#(3)} exhibits the most pronounced decline within this group, suggesting that the holistic level is particularly effective in disambiguating modification object pronouns, thereby enhancing the overall understanding of the multi-modal query. \textbf{D\#(3)} also shows the greatest performance drop among the three derivatives in this group on the CIR datasets, indicating that the proposed \textit{Holistic Pronoun Disambiguation} module is instrumental in facilitating composed semantic learning for the image modality. This demonstrates the module's strong generalization ability in handling image data.

\noindent
\textit{$\bullet$ \textbf{M2:Ablation on Atomistic Uncertainty Modeling.}} This group is designed to validate the effect of \textit{Atomistic Uncertainty Modeling} and obtain the following derivatives. \textbf{D\#(4)~w/o\_A\_Detail} targets to explore the effect of visual detail embedding via removing it. \textbf{D\#(5)~w/o\_A\_Compose} replaces the atomistic composition with simple addition to validate its impact. \textbf{D\#(6)~w/o\_Atomistic\_Level} ablates the whole \textit{Atomistic Uncertainty Modeling} and corresponding atomistic features to assess the effect of the atomistic level.

Based on the results of \textbf{\textit{M2}} in Table~\ref{tab:ablation}, we can draw the following conclusions. 
\textbf{1)} \textbf{D\#(4)} and \textbf{D\#(5)} exhibit performance degradation compared with the complete HUD model, which indicates that the \textit{Atomistic Uncertainty Modeling} indeed enables the model to discern the ``key details'' in the visual content, thus enhancing HUD's focus on fine-grained details and improving retrieval performance. They also exhibit performance degradation on the CIR datasets, which indicates that there is also a need to discern the ``key detail'' in the image modality. The proposed \textit{Atomistic Uncertainty Modeling} and atomistic composition effectively enhance HUD's focus on the fine-grained details in the image modality.
\textbf{2)} The performance drop observed in \textbf{D\#(6)} is more pronounced than that of \textbf{D\#(3)}, underscoring the necessity for the model to concentrate on ``key details''. The proposed \textit{Atomistic Uncertainty Modeling} not only addresses the model's limited focus on ``key details'' but also complements the holistic semantics, thereby further enhancing HUD's detail-focused capability. Moreover, \textbf{D\#(6)} shows a more pronounced drop in the fashion-domain compared to the open-domain dataset. This is reasonable because the ``key details'' in fashion-domain images typically represent only a small portion of the cloth (\textit{e.g.,} the collar or sleeves), necessitating atomistic-level interactions to ensure that the model accurately attends to these visually minor yet critical ``key details''.

\noindent
\textit{\textbf{$\bullet$ M3:Ablation on Holistic-to-Atomistic Alignment.}} This group aims to explore the effect of \textit{Holistic-to-Atomistic Alignment} and comprises the following derivatives. 
\textbf{D\#(7)  w/o\_Bias} intends to validate the impact of bias term $\mathcal{B}(\cdot,\cdot)$ via removing it in Eq.$($\ref{holistic_score}$)$.
\textbf{D\#(8) w/o\_$\mathcal{L}_{H\leftrightarrow A}$} and \textbf{D\#(9) w/o\_$\mathcal{L}_{rank}$} remove optimization functions separately in Eq.$($\ref{optimization}$)$ to explore their separate effect.

From the outcomes of \textbf{\textit{M3}} in Table~\ref{tab:ablation}, the following insights can be observed.
\textbf{1)} \textbf{D\#(7)} exhibits a performance decline, which highlights the necessity of eliminating the semantic bias introduced by the sampled \textit{i.i.d.} random variables in order to enhance the retrieval capability of the model. 
\textbf{2)} The performance of \textbf{D\#(8)} is inferior to that of HUD, indicating the necessity for distributional consistency between the holistic and atomistic levels. This consistency avoids conflicts between holistic and atomistic semantics, thereby improving the composed retrieval effectiveness at both levels on both CVR and CIR tasks.
\textbf{3)} \textbf{D\#(9)} performs worse than HUD, which indicates the necessity of employing a batch-based classification loss to align the multi-modal query and target video at both the holistic and atomistic levels. Furthermore, the loss function also effectively drives the multi-modal query representations closer to the target image features in the CIR task.

\subsection{Sensitivity Analysis}
\begin{figure}[h]
    \centering
    % \vspace{-1.0em}
    \vspace{-2pt}
	\includegraphics[width=\linewidth]{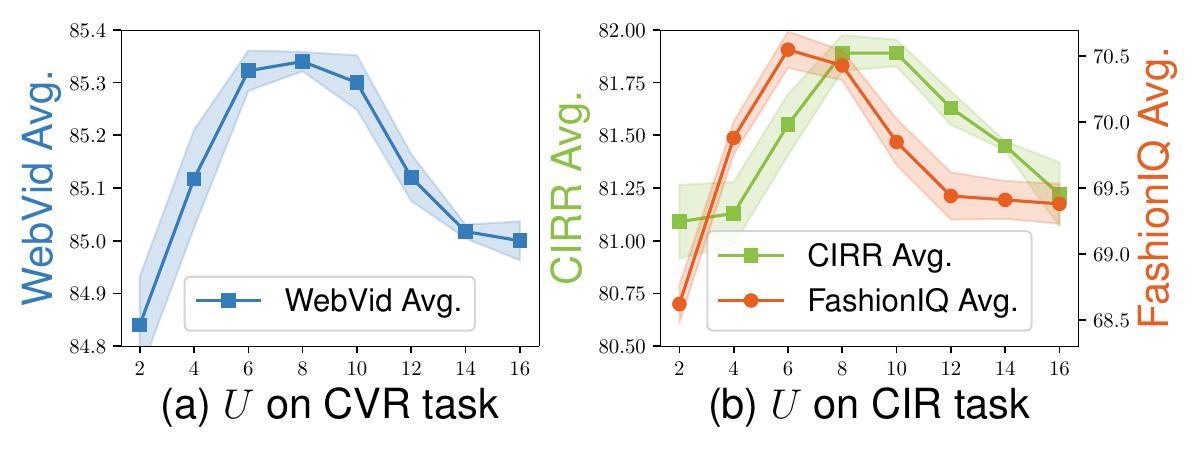}
      \vspace{-8pt}
	\caption{Sensitivity to Probabilistic Sample Number $U$ on (a) CVR and (b) CIR task.}
  \vspace{-5pt}
	\label{fig:num_U}
\end{figure}

Previously, the superior retrieval performance of HUD on both CVR and CIR tasks has been validated. In this section, we further explore the effectiveness of the probabilistic modeling employed by HUD through a sensitivity analysis on the probabilistic sample number $U$. 
As illustrated in Figure~\ref{fig:num_U}, we conduct a performance comparison under different sample numbers \(U\) (detailed in Eq.$($\ref{sample}$)$) on both the CVR and CIR tasks. The results shown in the figure indicate that as \(U\) increases, the performance of HUD on all four datasets initially improves and then decreases. This is reasonable since a larger number of samples is capable of modeling richer modification object semantics to resolve the ambiguity caused by ambiguous pronouns. However, given that the number of modification objects is limited, an excessive number of samples may lead to biases in disambiguation learning, thereby resulting in a decline in retrieval performance.

\subsection{Case Study}
\begin{figure}[ht!]
    \centering
    \vspace{-12pt}
	\includegraphics[width=0.95\linewidth]{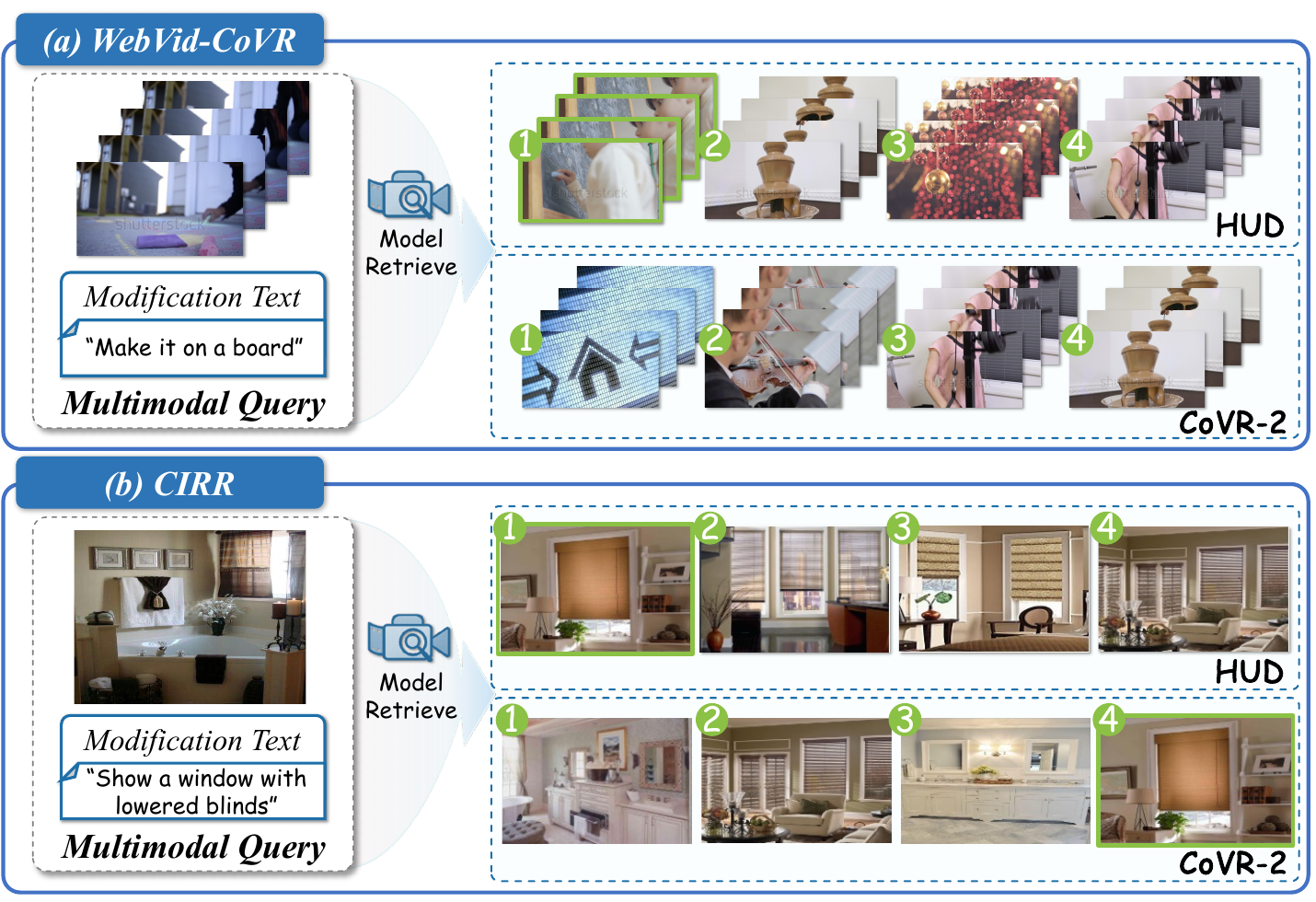}
\vspace{-12pt}
	\caption{Case study on (a) WebVid-CoVR and (b) CIRR.}
    \vspace{-18pt}
	\label{fig:case_study}
\end{figure}

To visually validate the retrieval effectiveness of HUD, we present in Figure~\ref{fig:case_study} the top-$4$ retrieval results obtained by our proposed HUD and the representative CVR model CoVR-2~\cite{covr-2} on both the CVR and CIR datasets. From the results, we obtain the following observations.
\textbf{1)} As illustrated in Figure~\ref{fig:case_study}(a), we observe that HUD successfully retrieves the target video at the top rank, whereas CoVR-2 fails. This indicates that, based on the modification text term ``board'', HUD can infer that the pronoun ``it'' refers to the ``chalk'' in the reference video. Additionally, since the ``chalk'' occupies only a small visual area in the reference video but qualifies as a ``key detail''. Thus, it is likely to be overlooked by conventional CVR models. This further validates the efficacy of HUD's pronoun disambiguation and detail uncertainty modeling.
\textbf{2)} In Figure~\ref{fig:case_study}(b), we observe that HUD places the target image at the top rank while CoVR-2 ranks it fourth. Moreover, the top four retrieval results returned by HUD align more closely with the expected requirements of the multi-modal query compared to those of CoVR-2. We also note that the ``window'' mentioned in the modification text occupies only a small region in the reference image (appearing only at the top right corner). This observation underscores the necessity of the \textit{Holistic-to-Atomistic Alignment} module within HUD, which is crucial for capturing fine-grained modification semantics. By modeling the atomistic-level compositional associations, HUD can more effectively focus on the ``window'' that needs to be modified, thereby achieving more accurate retrieval results.

\section{Conclusion}
In this work, we investigated the challenging CVR task. Although previous approaches achieved remarkable progress, they overlooked the disparity in information density between the video and text modalities. This limitation further resulted in two critical issues: (1) modification subject referring ambiguity, and (2) limited detailed semantic focus. To address these challenges, we proposed a novel CVR framework, termed the Hierarchical Uncertainty-aware Disambiguation network (HUD). HUD was the first framework to leverage the disparity in information density between video and text for improved multi-modal query understanding. By capturing overlapping semantics through holistic cross-modal interaction and modeling atomistic semantics, HUD effectively resolved modification subject referring ambiguity and enhanced attention to ``key detail'', thereby enabling accurate composed feature learning. Furthermore, our proposed HUD also generalized well to the Composed Image Retrieval (CIR) task and achieved state-of-the-art performance on three benchmark datasets across both CVR and CIR tasks. In future work, we plan to extend our method to multi-turn composed multi-modal retrieval.

\section*{Acknowledgments}
This work was supported in part by the National Natural Science Foundation of China, No.:62276155, No.:62476071, No.:U24A20328, and No.:624B2047; in part by the Guangdong Basic and Applied Basic Research Foundation, No.:2025A1515011732; in part by the China National University Student Innovation \& Entrepreneurship Development Program, No.:202410422071.
% \clearpage
% \balance

%%
%% The next two lines define the bibliography style to be used, and
%% the bibliography file.

\clearpage

% 1. 页面与计数器设置 (可选)
\onecolumn % 如果你想让标题跨栏居中（通常看起来更像独立标题）
\setcounter{page}{1} % 重置页码为 1
\renewcommand{\thepage}{S\arabic{page}} % 页码显示为 S1, S2...

% 2. 手动制作“大标题”

% 3. 切回双栏（如果需要保持 ACM 格式）
\twocolumn[{ 
  \centering
  \vspace*{1cm}
  
  % 标题部分：用 \par 换行，不要用 \\
  {\Huge \bfseries Supplementary Material \par} 
  
  \vspace{0.5cm}
  
  {\Large for \textit{\textbf{``HUD: Hierarchical Uncertainty-Aware Disambiguation Network for Composed Video Retrieval''}} \par}
  
  \vspace{1cm}
}]

% \begin{bibunit}[ACM-Reference-Format]
\appendix

\noindent
This is the supplementary material of the submitted paper \textit{\textbf{``HUD: Hierarchical Uncertainty-Aware Disambiguation Network for Composed Video Retrieval''}}. The content catalog is as follows:
\begin{itemize}[leftmargin=16pt]
    \item \textbf{Appendix~\ref{appendix:dataset_detailes}}: a detailed explanation of datasets.
    \item \textbf{Appendix~\ref{appendix:sensitivity}}: additional experiments about the sensitivity analysis to hyperparameter $\kappa$.
    \item \textbf{Appendix~\ref{appendix:probabilistic}}: probabilistic embedding visualization.
    \item \textbf{Appendix~\ref{appendix:cir}}: more retrieval examples for HUD on the four datasets.
\end{itemize}

\section{Details of Datasets}
\label{appendix:dataset_detailes}
For a comprehensive evaluation, we select four datasets, which include one CVR dataset, WebVid-CoVR~\cite{covr}, and two CIR datasets, FashionIQ~\cite{FashionIQ} and CIRR~\cite{cirr}. We now describe each dataset in detail as follows.

\subsection{CVR datasets}
We select the large-scale dataset WebVid-CoVR for the evaluation on the CVR task.
\begin{itemize}[leftmargin=16pt]
\item \textbf{WebVid-CoVR}~\cite{covr} serves as the first large-scale benchmark tailored for the CVR task. This dataset is generated by processing the WebVid-2M\cite{Webvid-2M} dataset, yielding \mbox{$\sim$\hspace{0em}$1.6$} million CVR triplets that span \mbox{$\sim$\hspace{0em}$131k$} unique videos and \mbox{$\sim$\hspace{0em}$467k$} distinct modification texts. On average, each video lasts around $16.8$ seconds, and each modification text consists of approximately $4.8$ words. Moreover, each target video is associated with roughly $12.7$ triplets. The test set of WebVid-CoVR includes $2.5k$ high-quality triplets, which were carefully chosen from the separate WebVid-10M~\cite{Webvid-2M} dataset after an intensive process of annotation and noise removal, resulting in a robust and challenging evaluation benchmark.
\end{itemize}

\subsection{CIR datasets}
We select the fashion-domain dataset, FashionIQ, and the open-domain dataset CIRR for the evaluation on the CIR task.

\begin{itemize}[leftmargin=16pt]
\item \textbf{FashionIQ}\cite{FashionIQ} is a widely used retrieval dataset in the fashion domain for CIR tasks, and the images are sourced from \textit{Amazon.com}. The dataset comprises $77,684$ images split into three categories: \textit{Dresses}, \textit{Shirts}, and \textit{Tops\&Tees}. Following the evaluation protocol outlined in~\cite{FashionIQ}, we treat each category as a separate dataset for our experiments. Consistent with previous CIR approaches, approximately $46k$ images are allocated for training and around $15k$ for testing, resulting in about $18k$ triplets for training and roughly $6k$ triplets for testing.

\item \textbf{CIRR}\cite{cirr} is an open-domain dataset introduced for the CIR task. It consists of $21,552$ real images sourced from the well-known language reasoning dataset 
${\operatorname{NLVR}}^2$~\cite{NLVR2}, which has been widely applied in natural language reasoning studies. The CIRR dataset comprises a total of \mbox{$\sim$\hspace{0em}$36.5K$} triplets, with $80\%$ allocated for training, $10\%$ for validation, and the remaining $10\%$ for testing. Additionally, CIRR features a specialized subset designed for fine discrimination. This subset includes negative images that are visually very similar, thereby facilitating the evaluation of the model's ability to distinguish between subtle false negatives.
\end{itemize}

\section{More Experiments on Sensitivity Analysis}
\label{appendix:sensitivity}
\begin{figure}[h]
\centering
%\vspace{0.5em}
% \begin{center}
\includegraphics[width=0.95\linewidth]{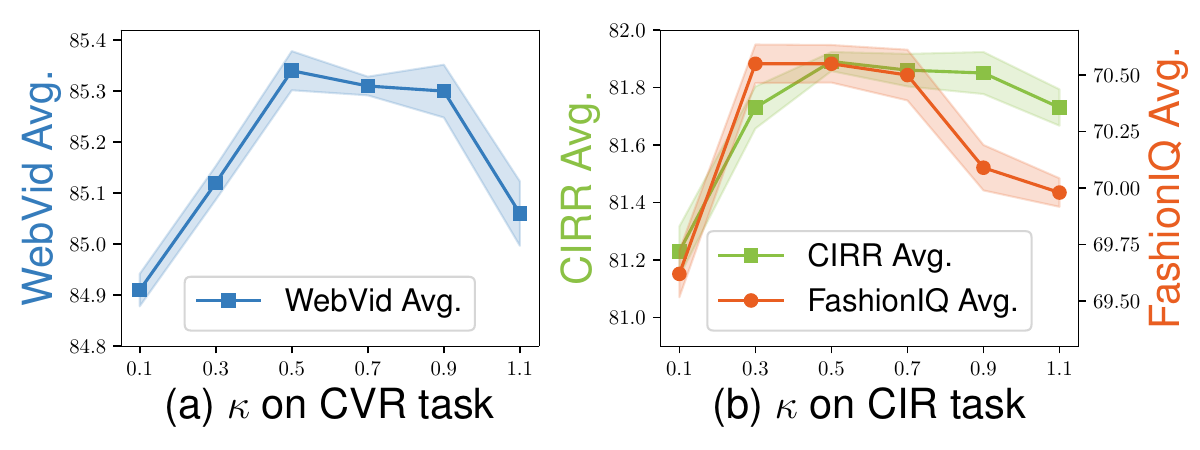}
  \vspace{-5pt}
% \end{center}
   \caption{Sensitivity to the trade-off hyper-parameter $\kappa$ on the (a) CVR task, and (b) CIR task.}
     \vspace{-8pt}
\label{fig:sensitive_kappa}
% \vspace{0.5em}
\end{figure}

To investigate the sensitivity of HUD to the hyperparameter \(\kappa\) in Eq.$($16$)$, we present in Figure~\ref{fig:sensitive_kappa} a performance comparison across different \(\kappa\) values on both the CVR and CIR tasks. 
The results allow us to draw the following conclusions.
As \(\kappa\) increases, the model's retrieval performance initially improves and then declines. Moreover, the performance fluctuations remain within a range of $1$. For example, on the WebVid-CoVR dataset, the difference between the highest and lowest performance values is only $0.43$. This indicates that HUD is robust with respect to the hyperparameter \(\kappa\), showing no significant performance variation within a certain range. 
However, an excessively high \(\kappa\) may lead to a slight performance decline, potentially because an overly large \(\kappa\) causes the modification objects attended by the holistic and atomistic distributions to converge, resulting in some semantic overlap and performance drop.

\section{Probabilistic Embedding Visualization}

\begin{figure}[h]
    \centering
    % \vspace{-1.0em}
	\includegraphics[width=1.0\linewidth]{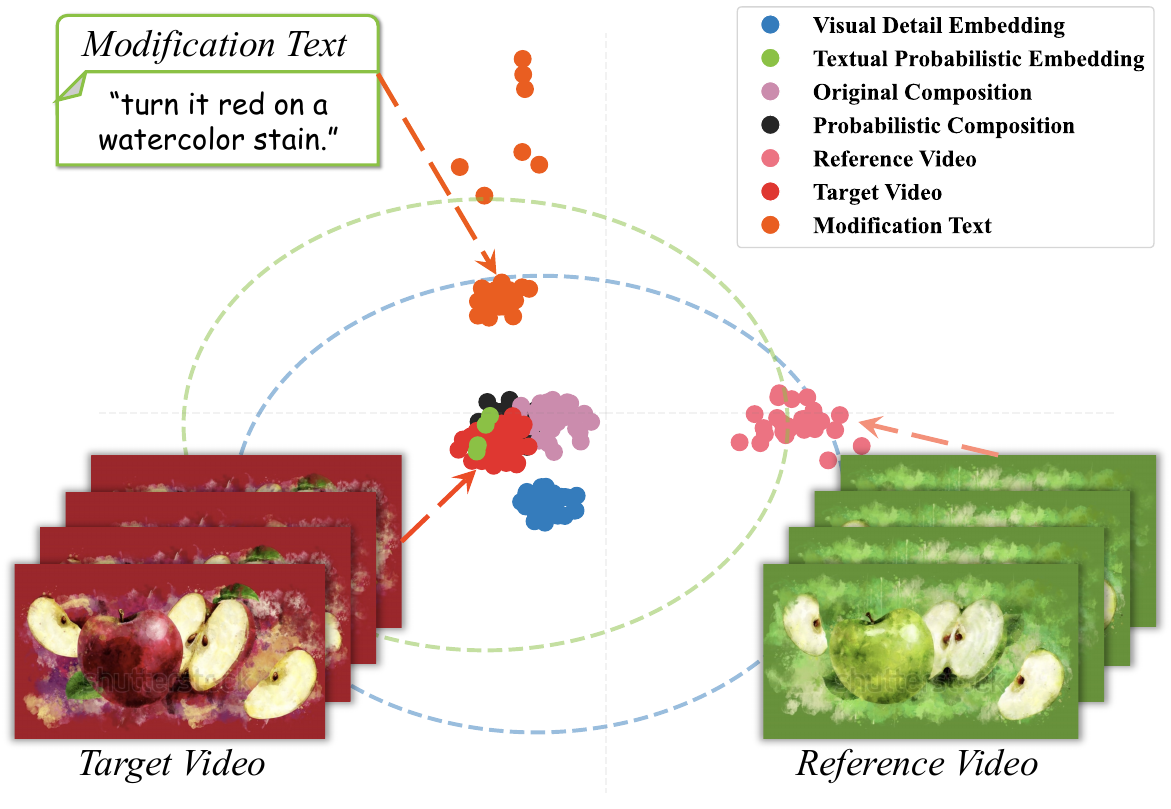}
    % \vspace{-2.5em}
	\caption{Visualization of the Probabilistic Embedding.}
    % \vspace{-1.2em}
	\label{fig:tsne}
\end{figure}

In order to intuitively analyze the role of the probabilistic embedding modeled by HUD, we employ t-SNE~\cite{tsne} to visualize the relative positions of various features in the embedding space, as shown in Figure~\ref{fig:tsne}. In the figure, different colored points represent the \textcolor[HTML]{357CBC}{\textit{visual detail embedding}}, \textcolor[HTML]{8BC145}{\textit{textual probabilistic embedding}}, \textcolor[HTML]{CB8CAD}{\textit{original composition}}, \textcolor[HTML]{262626}{\textit{probabilistic composition}}(with probabilistic embedding), as well as the \textcolor[HTML]{EC7382}{\textit{reference}}, \textcolor[HTML]{DF3731}{\textit{target}}, and \textcolor[HTML]{E95E21}{\textit{modification}} tokens. 
The \textcolor[HTML]{8BC145}{green} and \textcolor[HTML]{357CBC}{blue} ellipses represent the variance ranges for the textual and visual modalities, respectively. From the visualization, we obtain the following observations.

\label{appendix:probabilistic}

\textbf{1)} Regarding the visual variance, only a subset of modification tokens falls within the range, and for the textual variance, only a portion of reference tokens is encompassed. This suggests that during the processes of textual probabilistic embedding and visual detail embedding, the probabilistic modeling selectively focuses on the overlapping semantic components between the visual semantics and the modification semantics in the text, thereby achieving pronoun disambiguation and targeted focus on ``key details''.
\textbf{2)} Compared with the original reference and modification tokens, the visual detail embedding and textual probabilistic embedding are closer to the target tokens. This indicates that the probabilistic embedding, through cross-modal interactions capturing overlapping semantics, is capable of learning the modification-relevant semantic information so that both the visual and textual modalities accurately attend to the content that requires modification.
\textbf{3)} The semantic overlap between the probabilistic composition and the target tokens is significantly higher than that between the original composition and the target tokens. This demonstrates that leveraging probabilistic embedding in the compositional process more effectively comprehends the modification text and captures the targeted object within the visual domain, ultimately yielding more precise composed features.

\section{More CVR\&CIR Cases}
\label{appendix:cir}
\begin{figure*}[h]
\centering
%\vspace{0.5em}
% \begin{center}
\includegraphics[width=0.95\linewidth]{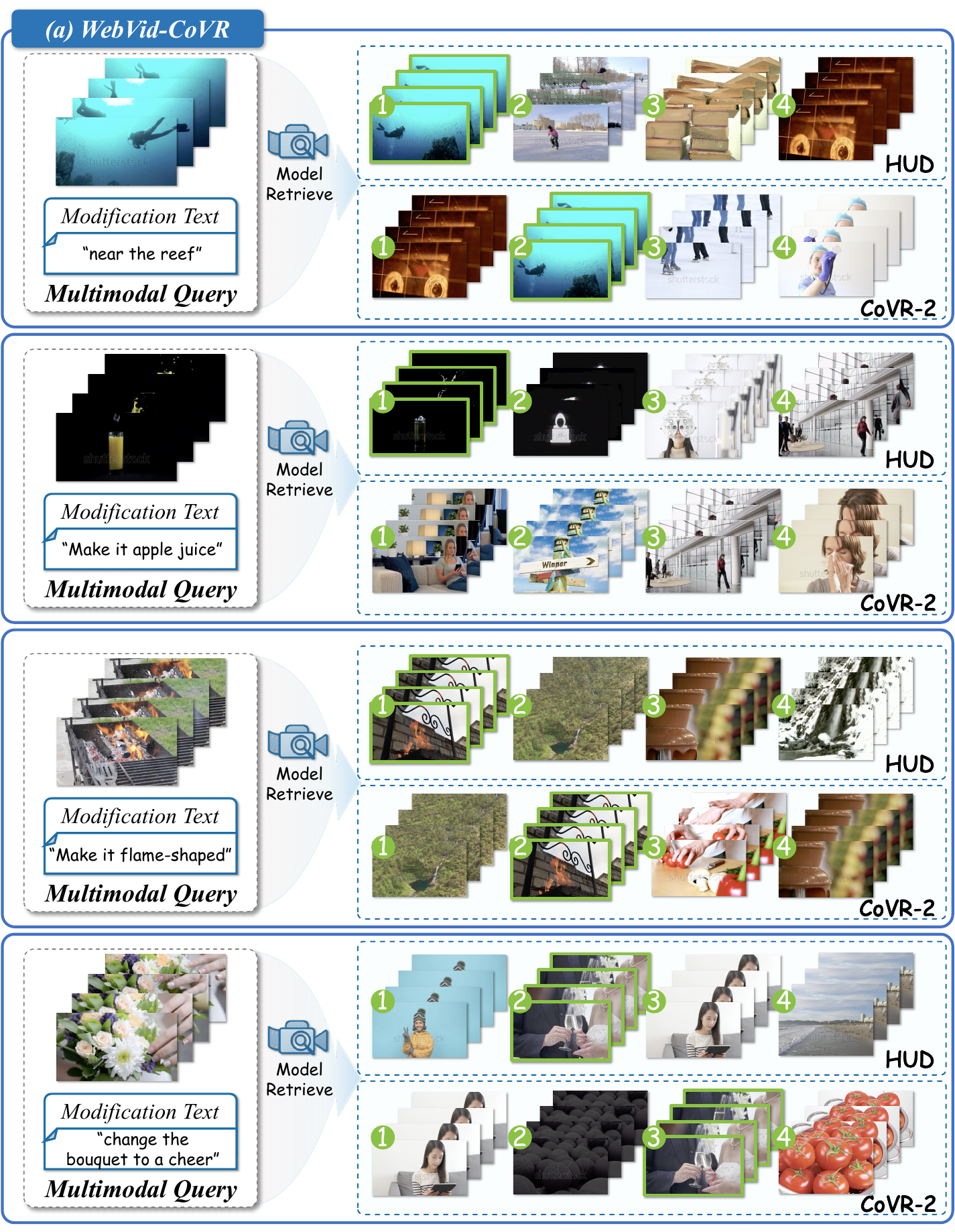}
  \vspace{-5pt}
% \end{center}
   \caption{Qualitative examples of HUD and CoVR-2 on the WebVid-CoVR dataset. The ground-truth target images are visualized by green boxes.}
     \vspace{-8pt}
\label{fig:case_webvid}
% \vspace{0.5em}
\end{figure*}

\begin{figure*}[h]
\centering
%\vspace{0.5em}
% \begin{center}
\includegraphics[width=0.95\linewidth]{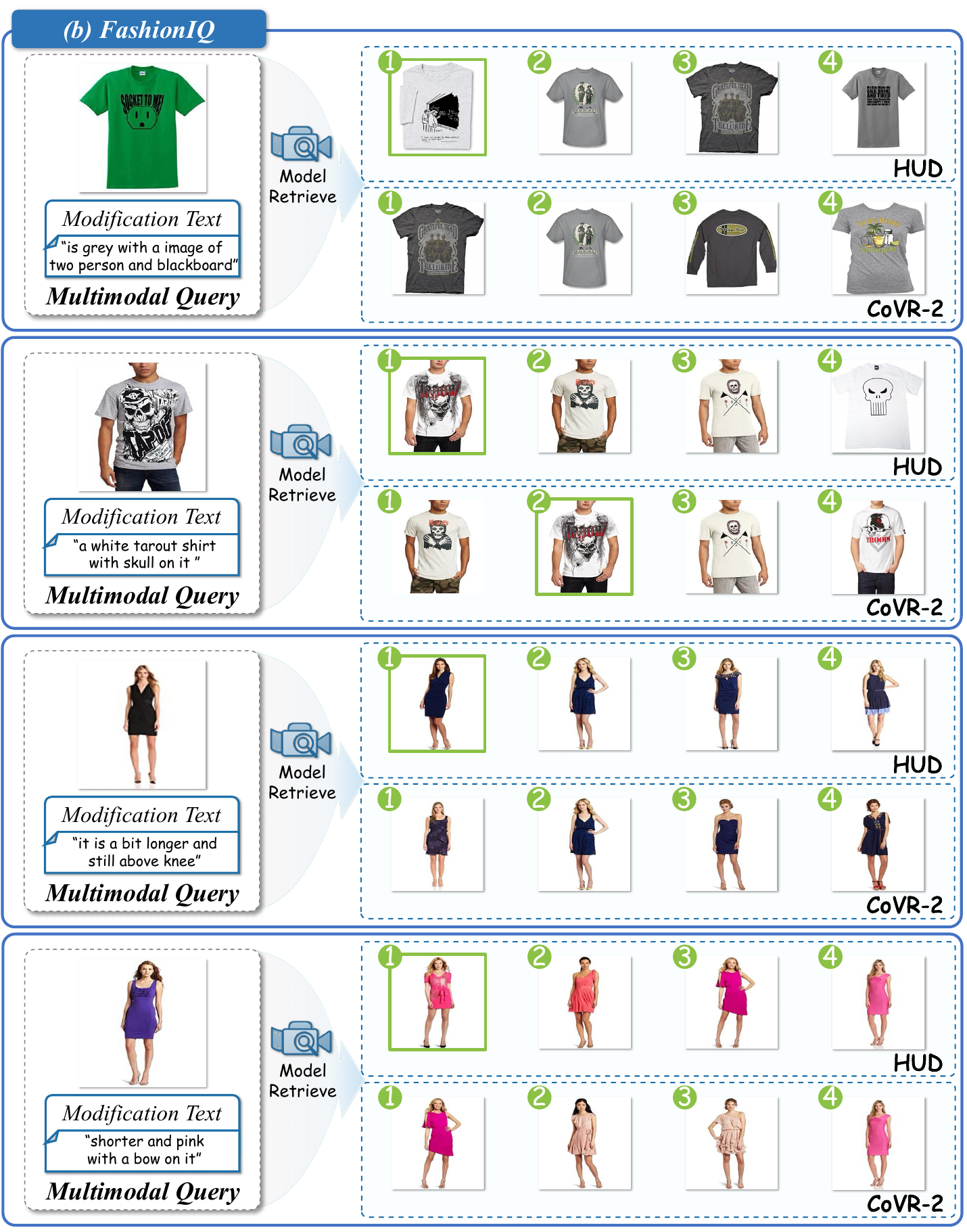}
  \vspace{-5pt}
% \end{center}
   \caption{Qualitative examples of HUD and CoVR-2 on the FashionIQ dataset. The ground-truth target images are visualized by green boxes.}
     \vspace{-8pt}
\label{fig:case_fashioniq}
% \vspace{0.5em}
\end{figure*}

\begin{figure*}[h]
\centering
%\vspace{0.5em}
% \begin{center}
\includegraphics[width=0.95\linewidth]{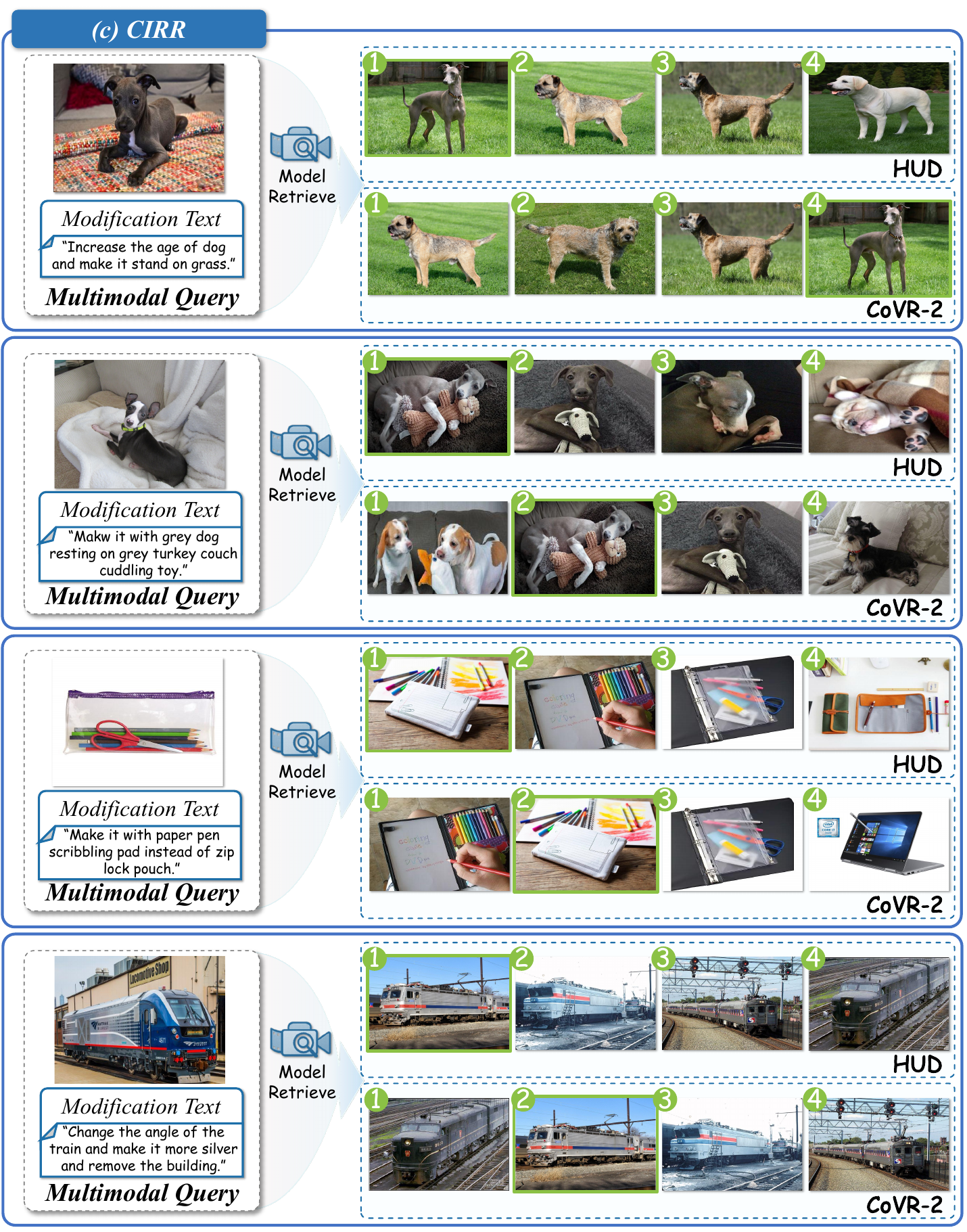}
  \vspace{-5pt}
% \end{center}
   \caption{Qualitative examples of HUD and CoVR-2 on the CIRR dataset. The ground-truth target images are visualized by green boxes.}
     \vspace{-8pt}
\label{fig:case_cirr}
% \vspace{0.5em}
\end{figure*}

As illustrated in Figure~\ref{fig:case_webvid}, Figure~\ref{fig:case_fashioniq}, and Figure~\ref{fig:case_cirr}, to provide a more comprehensive and intuitive demonstration of HUD's superior retrieval performance, in this section we further present additional CVR cases on the WebVid-CoVR datasets, as well as CIR cases on the FashionIQ and CIRR datasets. Each figure includes multi-modal queries, which comprises a reference video/image and a modification text, as well as the top-$4$ retrieval results retrieved by HUD and the most representative CVR model on the corresponding dataset. The ground truths (target videos or images) are indicated by \textcolor[HTML]{8BC145}{green} boxes. 
As the figures demonstrate, HUD is capable of ranking the target video or image at higher ranks in the retrieval lists, which indicates that leveraging the disparity in information density between modalities can indeed enhance retrieval performance.

Furthermore, in the last case presented in Figure~\ref{fig:case_webvid}, we showcase a failure case of the model. For example, as shown in the visualization, HUD ranks the ground truth video in the second position. However, the modification text ``change the bouquet to a cheer'' is semantically ambiguous, as ``cheer'' can refer to either a celebratory emotion or the specific act of toasting. In response, the top video retrieved by HUD depicts a woman making a happy, celebratory gesture, which aligns with the literal interpretation of ``cheering'', while the second video captures the scene of "toasting" with glasses. Therefore, the retrieved results by HUD actually satisfy the requirements of the multi-modal query, which further demonstrates that HUD effectively comprehends the modification instructions even when the textual query involves lexical ambiguity.

\section{More Related Work}
\textbf{Large Language Models for Multimodal Retrieval}.
The advent of Large Language Models~\cite{xin2025lumina,wang2024enhancing,zhou2024human,chu2025dynamic,zhang2025rearank,zeng2025bridging,xin2025luminamgpt} (LLMs) has significantly revolutionized the field of vision-language understanding~\cite{qian2025adaptive,liu2024rag,zhang2025assessing,liu2025qfft,qian2025dyncim,lu2025journey,zhou2025glimpse}. In the context of retrieval tasks, LLMs contribute primarily by enhancing the comprehension of the query text. For Composed Video Retrieval (CVR), the modification text often involves complex logic (e.g., ``change object A to B but keep the background C''). LLM-based architectures~\cite{shen2025expertflow,chu2025selective,chu2025mcam,zhang2024seppo,li2024synergized,xin2024v} can leverage their immense pre-trained knowledge to perform reasoning on these instructions~\cite{luo2025tr,sha2024hierarchical,kong2024tiger,kong2025accelerating}, thereby extracting richer query representations~\cite{11197022,sha2022regional,li2025hyfacialhybridfeatureextraction,yi2024towards}. 

However, deploying full-scale LLMs for the retrieval of fine-grained visual details remains challenging, particularly in distributed or privacy-sensitive scenarios such as \textbf{Federated Learning (FL)}~\cite{liu2025consistency}. First, the sheer model size and inference latency of LLMs are often prohibitive for FL settings~\cite{liuimproving}, where models must be lightweight enough to run on resource-constrained edge devices while minimizing communication overhead~\cite{zhao2025recondreamer++, xin2024vmt,xin2024mmap,xin2024parameter,lu2025differentiable}. Second, while LLMs excel at general semantic reasoning~\cite{xiang2025promptsculptor,liu2025fedadamw,xu2025mmm,zhang2024building,lu2025causalsr,liu2024fedbcgd}, they may struggle to adapt to the \textit{statistical heterogeneity} of local client data or lack the granularity required to ground ``atomistic'' visual details (such as the small ``chalk'' or ``window'' instances discussed in our case studies) without centralized aggregation. Our HUD framework addresses these limitations through a lightweight hierarchical design. Instead of relying on the heavy computation of end-to-end Video-LLMs, HUD explicitly models uncertainty and incorporates an efficient atomistic-level alignment module. This ensures that the model captures subtle visual modifications and is amenable to deployment in decentralized environments, balancing precise retrieval with the efficiency required for potential federated applications.

% \putbib[reference]
% \bibliographystyle{ACM-Reference-Format}
% \bibliography{reference}
% \end{bibunit}
%% End of file `sample-sigconf.tex'.

\bibliographystyle{ACM-Reference-Format}
\bibliography{reference}

% \end{sloppypar}
% \end{CJK}
\end{document}